\documentclass[lettersize,journal]{IEEEtran}
\usepackage{amsmath,amsfonts}
\usepackage{algorithmic}
\usepackage{algorithm}
\usepackage{array}
\usepackage[caption=false,font=normalsize,labelfont=sf,textfont=sf]{subfig}
\usepackage{textcomp}
\usepackage{stfloats}
\usepackage{url}
\usepackage{verbatim}
\usepackage{graphicx}
\usepackage{cite}
\usepackage{multirow}
\usepackage{subfig}
\usepackage{graphicx}
\usepackage[table,xcdraw]{xcolor}
\usepackage{mathabx}
\usepackage{booktabs}

\begin{document}

\title{PRO-Face S: Privacy-preserving Reversible Obfuscation of Face Images via Secure Flow}

\author{
Lin~Yuan, Kai~Liang, Xiao~Pu, Yan~Zhang, Jiaxu~Leng, Tao~Wu,~\IEEEmembership{Member,~IEEE}, Nannan~Wang,~\IEEEmembership{Member,~IEEE}, and Xinbo~Gao,~\IEEEmembership{Senior Member,~IEEE}
\thanks{This work is supported by the National Natural Science Foundation of China under Grant No. 62201107, 62036007 and U22A2096, in part by the Natural Science Foundation of Chongqing under Grand No. CSTB2022NSCQ-MSX1265 and CSTB2022NSCQ-MSX1342, in part by the Chongqing Excellent Scientist Project under Grant No. cstc2021ycjh-bgzxm0339, and in part by the Creative Research Groups of Chongqing Municipal Education Commission under Grant CXQT21020. (\textit{Corresponding author: Xinbo Gao}.)}
\thanks{Lin Yuan, Xiao Pu, and Tao Wu are with the School of Cyber Security and Information Law, Chongqing University of Posts and Telecommunications, Chongqing 400065, China (e-mail: yuanlin@cqupt.edu.cn; puxiao@cqupt.edu.cn; wutao@cqupt.edu.cn)}
\thanks{Kai Liang and Nannan Wang are with the School of Optoelectronic Engineering, Chongqing University of Posts and Telecommunications, Chongqing 400065, China; Nannan Wang is also with the School of Telecommunications Engineering, Xidian University, Xi'an 710071, China (e-mail: kailiang777@qq.com; nnwang@xidian.edu.cn)}
\thanks{Yan~Zhang, Jiaxu~Leng and Xinbo~Gao are with the School of Computer Science and Technology, Chongqing University of Posts and Telecommunications, Chongqing 400065, China (e-mail: yanzhang1991@cqupt.edu.cn; lengjx@cqupt.edu.cn; gaoxb@cqupt.edu.cn)}
}

\markboth{Journal of \LaTeX\ Class Files,~Vol.~14, No.~8, August~2021}%
{Shell \MakeLowercase{\textit{et al.}}: A Sample Article Using IEEEtran.cls for IEEE Journals}

\IEEEpubid{0000--0000/00\$00.00~\copyright~2021 IEEE}

\maketitle

\begin{abstract}
%
  This paper proposes a novel
  paradigm for facial privacy protection that unifies 
  multiple characteristics including anonymity, 
  diversity, reversibility and security within a 
  single lightweight framework.
  We name it \textbf{PRO-Face S}, short for
  \textbf{P}rivacy-preserving \textbf{R}eversible 
  \textbf{O}bfuscation of \textbf{Face} images via 
  \textbf{S}ecure flow-based model.
  In the framework, an Invertible Neural Network (INN) is utilized 
  to process the input image along with its pre-obfuscated form, 
  and generate the privacy protected 
  image that visually approximates to the pre-obfuscated one, 
  thus ensuring privacy.
  The pre-obfuscation applied can be in diversified form 
  with different strengths and styles specified by users. 
  Along protection, a secret key is injected into the network
  such that the original image can only be recovered from the protection image 
  via the same model given the correct key provided.
  Two modes of image recovery are devised to deal with
  malicious recovery attempts in different scenarios.
  Finally, extensive experiments conducted on three public image datasets 
  demonstrate the superiority of the proposed framework over 
  multiple state-of-the-art approaches.
\end{abstract}

\begin{IEEEkeywords}
facial privacy protection, image obfuscation, invertible neural network.
\end{IEEEkeywords}

\section{Introduction} \label{sec:intro}
\IEEEPARstart{A}{dvancements} of video capturing and analytic 
technologies have brought tremendous convenience to public, 
but on the negative side, have also raised 
growing concerns over privacy.
Personal face photos being captured and published 
(intentionally or unintentionally) 
are prone to being abused by unauthorized parties, 
exposing great threats to individuals privacy and security. 
The case of ClearView AI~\cite{clearviewai} collecting personal 
photos and training facial recognition systems 
without users consent raised the alert on 
hidden privacy threats behind public face images. 
Breach of surveillance cameras~\cite{breachcamera} 
implies tremendous privacy and security 
risks underneath video surveillance systems, 
especially when not being properly regulated or secured. 
China's first lawsuit on facial recognition~\cite{cgnt} 
initiated by Professor Guo in Hangzhou also drew public 
attention towards the legality and privacy of 
ubiquitous face recognition systems.

In academia, facial privacy protection methodologies 
(aka anonymization, de-identification, or DeID) 
have been extensively studied to deal with the privacy threats 
induced by either unauthorized machine recognition or unexpected human inspection. 
Early approaches utilize traditional image processing techniques
to distort facial appearance in image, such as 
filtering~\cite{erdelyi2014cartoon}, 
masking~\cite{yuan2017iet},
various transformation~\cite{korshunov2013warping}. 
With advancements of deep learning 
especially various generative models, 
diversified methods have been proposed to 
anonymize the facial appearance in image such that 
the the protected face looks differently from the original.
Different groups of solutions feature 
distinctive characteristics and advantages.
Conventional approaches are 
widely used in practice due to its simplicity 
and strong obfuscation effect to human, 
but are considered to be less effective 
in protecting privacy against machine vision.  
Generative approaches offer novel measures 
to hide the identity information  
while preserving the natural facial appearance. 
Although seemly promising, this type of approaches 
are still barely applied in practice due to ongoing 
arguments and potential legal issues related to generative fake media. 
{\it Hence, we are motivated to design a unified methodology that 
can assemble multiple obfuscations into one to provide users 
with full flexibility in choosing the desired one 
for fitting different applications. }
\IEEEpubidadjcol 

In this paper, 
we stand on a fresh view 
and propose a novel paradigm for facial privacy protection 
that unifies multiple excellent characteristics at once.
We name it \textbf{PRO-Face~S}, short for 
\textbf{P}rivacy-preserving 
\textbf{R}eversible 
\textbf{O}bfuscation of 
\textbf{Face} images via a
\textbf{S}ecure flow-based model.
In such a framework (as illustrated in Fig.~\ref{fig:illustration}),
the input image to be protected is first pre-obfuscated 
by a privacy-preserving image operation.  
Then, both the original and the pre-obfuscated images are fed into 
an invertible flow-based network, generating the protection image 
that visually approximates the pre-obfuscated one. 
Relying on the intrinsic network invertibility, 
the original image can be recovered with high fidelity 
using the same set of model parameters as protection,
in need of visual inspection or forensics.
Specially, the protection network is conditioned on 
a user-specified secret such that the correct recovery 
can only be achieved when the correct secret is available. 
While, any malicious recovery attempt using a wrong secret key 
will only produce distorted or obfuscated recovery images. 

Therefore, our framework elegantly unties multiple merits an ideal 
visual privacy protection method should have:
\begin{itemize}
	\item \textbf{Diversity:} Unlike existing approaches 
where ``diversity'' usually indicates diverse facial identities corresponding 
to different secret keys, we redefine the term to a broader notion 
indicating diverse types of visual obfuscations to support wider use cases. 	
	\item \textbf{Anonymity:} Due to the diversity in visual obfuscations to be applied, 
the framework offers different degrees of personalized anonymity 
depending on the type and strength of user-specific obfuscation.
	\item \textbf{Reversibility:} Thanks to the intrinsic invertibility of the flow-based model, 
the framework can easily achieve high quality image recovery with 
reduced number of model parameters, less training efforts, and better interpretability.
	\item \textbf{Security:} Relying on a specially designed keying mechanism, 
the protection and recovery processes are mutually controlled by a secret key, 
which ensures correct recovery only when the correct secret key is available. 
	\item \textbf{Lightweight:} Last but not least, 
the framework encapsulates a limited number 
of invertible blocks with less than 1M parameters to achieve optimal 
protection and recovery performances, 
which make it applicable for more practical applications.
\end{itemize}

This paper makes the following contributions:
\begin{enumerate}
	\item A novel {\it one-size-fits-all} paradigm for 
	facial privacy protection is proposed, which integrates 
	multiple merits including diversity, anonymity, reversibility 
	and security into a single lightweight framework.
	\item A novel key-based mechanism controlling forward (protection) 
	and backward (recovery) passes of flow-based model is devised, 
	which ensures secure image protection and high quality recovery simultaneously. 
	\item Extensive experiments and comprehensive analysis on privacy, 
	reversibility and security of the proposed framework are presented. 
	In addition, a small set of preprocessed 3K images is provided as 
	a standard benchmark for fair comparison in future research of the area. 
\end{enumerate}


\section{Related Work} \label{sec:related_work}
This section reviews the literatures that are most relevant to ours, 
and points out our inspirations from existing studies. 

\subsection{Image Processing Approaches}
Early research towards facial privacy protection 
mainly uses conventional image processing  
to conceal sensitive information (such as face) 
from input images~\cite{meden2021bpetsurvey}. 
Proposed approaches can be roughly categorized into 
masking-based methods~\cite{yuan2017iet}, 
filtering-based methods~\cite{erdelyi2014cartoon,zhou2021pixelate}, 
and transformation-based methods~\cite{korshunov2013warping,
ciftci2018falsecolor}. 
Different methods offer different degrees and styles of 
visual obfuscation. 
Although considered to offer less utility and limited reversibility, 
this group of approaches are still popularly used in practice, 
due to their simplicity and strong obfuscation effect.

\subsection{Deep-based Facial Anonymization}
The success of deep learning technologies has greatly 
boosted the research of face privacy protection 
towards diversified directions. 
A group of adversarial-based approaches 
~\cite{mirjalili2020privacynet,oh2017advimage,shan2020fawkes,hu2022cvpr}
have proven their success in fighting against unauthorized face recognition.  
However, they do not mean to protect image privacy against human inspection. 
Another group of approaches utilize generative models 
to create natural facial appearance different from original.
Early studies in this group include 
head inpainting~\cite{sun2018headinpainting}, 
DeepPrivacy~\cite{hukkelas2019deepprivacy},
live face de-identification~\cite{gafni2019iccv}, 
and CIAGAN~\cite{maximov2020ciagan}. 
Recent efforts have been devoted to 
more sophisticated solutions covering 
multiple utility properties such as 
diversity, reversibility, security, and identifiability.
Gu et al.~\cite{gu2020eccv} first proposes 
reversible face identity transformer that 
performs face anonymization and de-anonymization upon a binary password.
Cao et al.~\cite{cao2021iccv} proposes a personalized invertible DeID
framework where protection and recovery are implemented by 
transforming disentangled ID vector with user-specified password.
Wu et al.~\cite{wu2021pecam} designs PECAM, which performs privacy-enhanced 
securely-reversible video transformation in video streaming and analytics systems.
Li et al.~\cite{li2021acmmm,li2023tifs} 
proposes identity-preserved facial anonymization via 
identity-aware region discovery to determine facial attributes sensitive to human eyes.
Hugo et al.~\cite{hugo2022uunet} proposes UU-Net, 
a reversible face DeID framework powered by two U-Nets for protection 
and recovery respectively, to be in a photorealistic and seamless way 
in video surveillance.
Yuan et al.~\cite{yuan2022ivfg} devises an 
Identifiable Virtual Face Generator (IVFG) 
to produce the virtual faces that are 
recognizable via their virtual ID embeddings. 
Wen et al.~\cite{wen2022idmask} attempts to protect facial privacy 
in video frames via a modular architecture named IdentityMask, 
which leverages deep motion flow and protective motion IdentityMask, 
Zhang et al.~\cite{zhang2023rapp} proposes RAPP, 
a reversible privacy-preserving scheme for protecting 
various facial attributes. 
Recently, Li et al.~\cite{li2023riddle} proposes RiDDLE, 
a reversible DeID framework with StyleGAN~\cite{karras2019stylegan}
latent space encryption.

\subsection{Template-based Face Anonymization}
A small group of research utilizes a template-based methodology 
to generate the anonymized face that approximates a pre-obfuscated template. 
You et al.~\cite{you2021rppr} proposes a 
reversible privacy-preserving recognition 
framework that hides original image into its pixelated form, 
which can be later recovered based on a different network. 
Yuan et al.~\cite{yuan2022proface} introduces PRO-Face, 
which not only supports diverse facial obfuscations, 
but also preserves identity information in the protected faces. 
However, PRO-Face does not take into account reversibility by design.
Recently, Yang et al.~\cite{yang2023imn} first utilizes 
invertible neural network (INN) to build an 
invertible mask network (IMN) to achieve 
the state-of-the-art protection and recovery performance 
in the category of research. 
However, no security property has been considered 
in the IMN framework. 
The design of our method is highly inspired by 
PRO-Face~\cite{yuan2022proface} and IMN~\cite{yang2023imn}, 
whereas multiple usable features are integrated into one framework 
with optimal performance achieved on each aspect.

\subsection{Invertible Neural Networks}
Invertible neural network (INN) is a special network structure 
that employs a flow of bijective mapping functions to 
perform invertible transformation between inputs and outputs. 
The inputs to the INN model can be easily recovered 
from the outputs using the same model parameters by 
inverting the bijective transformations.
INN was first proposed by Dinh et al.~\cite{dinh2014nice} 
as Non-linear Independent Components Estimation (NICE) 
to model complex probability densities. 
It was later adapted to handle image processing tasks 
by Dinh et al.~\cite{dinh2017realnvp} using convolutional coupling layers. 
Later, Kingma et al.~\cite{kingma2018glow} proposes Glow, 
a generative flow with invertible 1$\times$1 convolutions, 
for more efficient and realistic image generation. 
Due to the excellent performance, 
INN has been used in many image-related tasks including 
image-to-image translation~\cite{quderaa2019revgan, ardizzone2021cinn}, 
image scaling~\cite{xiao2020iir},
super-resolution~\cite{lugmayr2020srflow},
denoising~\cite{liu2021denoiseinn},
compression~\cite{xie2021compressioninn},
steganography~\cite{jing2021hinet,guan2023deepmih}, 
and face anonymization~\cite{yang2023imn}.

Specifically, our study is greatly inspired 
by HiNet~\cite{jing2021hinet} and DeepMIH~\cite{guan2023deepmih}, 
two invertible image steganography 
frameworks with excellent concealment and recovery performance. 
Similar as PRO-Face~\cite{yuan2022proface} and IMN~\cite{yang2023imn}, 
we also simulate the privacy protection process as image steganography, 
where the original private information is concealed into a pre-obfuscated template, 
which can be later recovered in case of need. 
Built on the similar INN structure of~\cite{guan2023deepmih},
we make elaborate security enforcement via a newly designed 
keying mechanism to ensure secure privacy protection and recovery. 
Moreover, we greatly reduce model parameters to make it 
more applicable to practical usages without sacrificing much performance.

\section{Problem Formulation} \label{sec:formulation}
\begin{figure}[t]
	\includegraphics[width=\columnwidth]{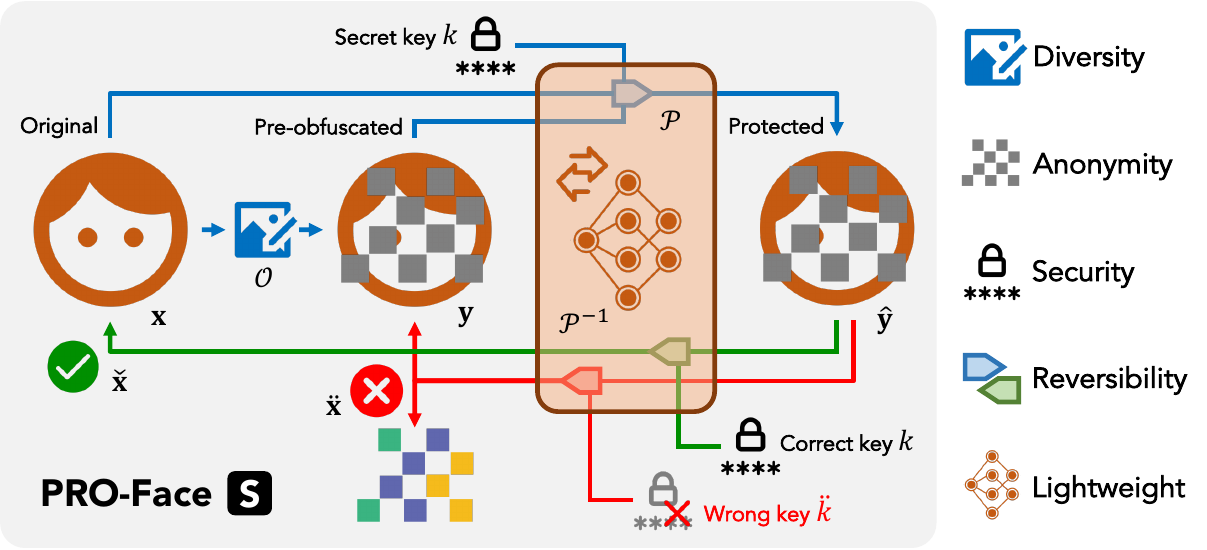}
	\caption{Illustration of the PRO-Face S paradigm for facial privacy protection.}
	\label{fig:illustration}
\end{figure}

We first formulate the proposed privacy protection paradigm (illustrated in Fig.~\ref{fig:illustration}), 
consisting of three major steps: pre-obfuscation, key-based protection, 
and secure recovery.  

\subsection{Pre-Obfuscation} 
Given an input image $\mathbf{x} \in \mathbb{R}^{3 \times W \times H}$, 
it is first processed by an existing image obfuscator  
$\mathcal{O}:~\mathbb{R}^{3 \times W \times H} \mapsto \mathbb{R}^{3 \times W \times H}$, 
resulting in the pre-obfuscated image $\mathbf{y} \in \mathbb{R}^{3 \times W \times H}$ 
that is significantly different from its original in terms of visual perception 
(noted by $\mathcal{V}$):
\begin{equation}
	\mathbf{y} = \mathcal{O}(\mathbf{x}), \ \mathcal{V}(\mathbf{y}) \ncong \mathcal{V}(\mathbf{x}).
\end{equation}
The pre-obfuscated image is assumed to have visual privacy removed.
Notably, the operation $\mathcal{O}$ can be in principle any type of facial obfuscator 
in diversified and personalized form. 
Thus, the pre-obfuscator in the framework offers high diversity 
in terms of obfuscation effect, areas and strengths. 

\subsection{Key-based Protection} 
As the second step, both the original and 
the pre-obfuscated image $(\mathbf{x}$, $\mathbf{y})$ 
are fed into an invertible protection module $\mathcal{P}$, producing the 
protection image $\mathbf{\hat{y}} \in \mathbb{R}^{3 \times W \times H}$ 
that is visually approximate to the pre-obfuscated one 
with high fidelity, and therefore being privacy-preserved: 
\begin{equation}
	\mathbf{\hat{y}} = \mathcal{P}(\mathbf{x}, \mathbf{y}, k), 
	\ \mathcal{V}(\mathbf{\hat{y}}) \cong \mathcal{V}(\mathbf{y}).
\end{equation}
Herein, a user-specified secret key $k$ is injected into the protection 
network to condition the generation of the protection image.
The role of the secret key is to control the mutual processes of 
protection and recovery such that the original face can only be 
recovered with the correct secret key.

\subsection{Secure Recovery}  
It is possible to recover the original image 
from the protection image with high fidelity 
given the correct secret key $k$ provided, 
by inverting the protection network (noted $\mathcal{P}^{-1}$) 
relying on the same model parameters:
\begin{equation}
	\check{\mathbf{x}} = \mathcal{P}^{-1}(\mathbf{\hat{y}}, k), \ 
	\mathcal{V}(\check{\mathbf{x}}) \cong \mathcal{V}(\mathbf{x}).
\end{equation}
Using any wrong secret key $\ddot{k}$ with even one bit 
difference from the original key, 
a wrong recovery image $\ddot{\mathbf{x}}$ will be resulted, 
which is significantly different from the original 
and fails to reveal any private visual information:
\begin{equation}
	\ddot{\mathbf{x}} = \mathcal{P}^{-1}(\mathbf{\hat{y}}, \ddot{k}), \ 
	\mathcal{V}(\ddot{\mathbf{x}}) \ncong \mathcal{V}(\mathbf{x})
\end{equation}
Such a key-controlled image recovery process 
features the security of our framework.

Interestingly, the protection and recovery processes roughly 
simulate a cycle of symmetric cryptography,
whereas the protection image in our case
(corresponding to the ciphertext) 
still preserves meaningful semantics 
instead of being completely randomized 
as cryptography.

\begin{figure*}[t]
\centering
	\includegraphics[width=\textwidth]{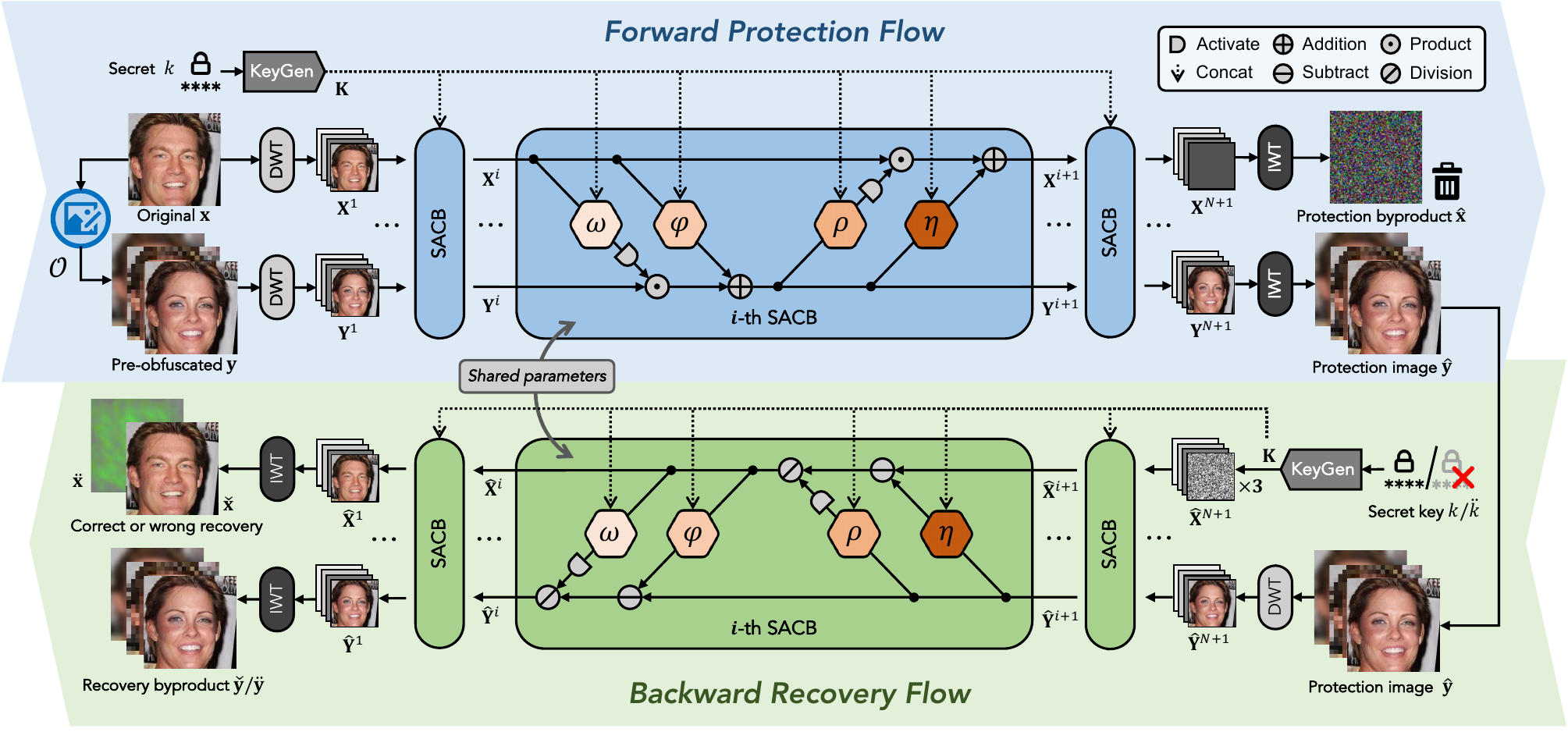}
	\caption{The architecture of PRO-Face S composed of the forward protection and backward recovery flows featured by several core components: pre-obufscation~($\mathcal{O}$), key generation (KeyGen), discrete wavelet transform and its inverse (DWT/IWT), and secure affine coupling block (SACB). }
	\label{fig:network_design}
\end{figure*}

\section{The Framework Design} \label{sec:method}
The workflow of the PRO-Face S framework is illustrated in 
Figure~\ref{fig:network_design},
the core of which is an invertible protection network 
composed of a flow of Secure Affine Coupling Blocks (SACBs). 
We adopt the basic design of the affine coupling block (ACB) 
structure from~\cite{guan2023deepmih}, 
and make innovative security enforcement on it by integrating 
randomness into each ACB conditioned on secret keys. 
In addition, Discrete Wavelet Transform (DWT) and 
Inverse Wavelet Transform (IWT) modules are also employed 
following~\cite{guan2023deepmih} to process input and output data, 
to better fuse the original image into the obfuscated template in the frequency domain, 
and to reduce the computational cost.
The distinctive design of our key generation module, 
and the image protection/recovery processes are described 
in detail as follows.

\subsection{Secret Key Generation}  
\label{sec:secret_key_generation}
\begin{figure}[t]
\centering
	\includegraphics[width=0.95\columnwidth]{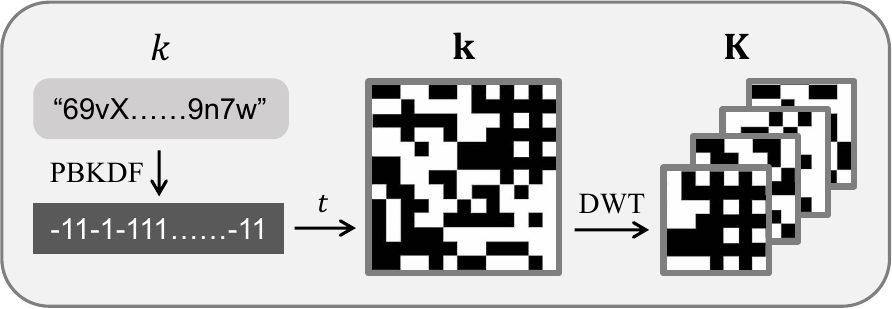}
	\caption{Illustration of the KeyGen module: 
	the initial secret $k$ in arbitrary length 
	is eventually mapped into a random secret map in shape 
	$(4 \times \frac{W}{2} \times \frac{H}{2})$.}
	\label{fig:keygen}
\end{figure}

The key generation module (KeyGen for short) 
aims to generate randomized information (named secret map) used to 
control the protection/recovery processes
based on an user-specified secret input. 
Simulating cryptography, 
we expect any difference in bits of the input key 
would result in totally different output secret map. 
Illustrated by Fig.~\ref{fig:keygen}, the KeyGen module works as follows:

Given an initial key $k$ with arbitrary characters and length 
(also known as password, to be specified by users), 
the Password-Based Key Derivation Function (PBKDF) 
is applied to derive from the initial key
a binary code with fixed length of $W\times H$,  
which is then geometrically transformed into a 2D binary map $\mathbf{k}$ 
with the same dimension as the input image (single channel):
\begin{equation} \label{eq:pbkdf}
	\mathbf{k} = t\left(\mbox{PBKDF}(k)\right), \ \mathbf{k} \in \{-1, 1\}^{W \times H}.
\end{equation}
The binary map $\mathbf{k}$ is further transformed into a 
four-channel secret map by DWT:
\begin{equation} \label{eq:secret_key_map}
	\mathbf{K} = \mbox{DWT}(\mathbf{k}),
\end{equation}
where $\mathbf{K}$ is in dimension of $(4 \times \frac{W}{2} \times \frac{H}{2})$.
The secret map $\mathbf{K}$ is the actual information 
integrated into the protection and recovery network 
for controlling the two processes mutually. 
Specifically, the version 2 of PBKDF (namely PBKDF2) 
with Hash-based Message Authentication Code (HMAC) pseudorandom function 
is applied, using a fixed salt and a small number of iterations 
(10 in our case) for demonstration purpose. 
The pseudorandomness of PBKDF2 ensures to output 
completely random secret map $\mathbf{K}$ 
for any unique initial secret.

\subsection{Forward Protection Flow} 
The forward flow of privacy protection is illustrated in the 
upper part of Fig.~\ref{fig:network_design}.
In the protection pass, the input face $\mathbf{x}$
is first obfuscated by an existing image obfuscater $o()$, 
resulting in pre-obfuscated image $\mathbf{y} = o(\mathbf{x})$.
Then, both $\mathbf{x}$ and $\mathbf{y}$ are decomposed into 
wavelet sub-bands by DWT: 
\begin{align}
	\mathbf{X}^1 &= \mathrm{DWT}\left(\mathbf{x}\right), \\
	\mathbf{Y}^1 &= \mathrm{DWT}\left(\mathbf{y}\right).
\end{align}
which are further fed into $N$ sequential Secure ACBs (SACBs for short).

Similar as~\cite{guan2023deepmih}, each SACB consists of four 
non-linear mapping functions noted $\omega()$, $\varphi()$, $\rho()$ and $\eta()$ 
respectively.  
Those functions share the same densely connected conv 
structure like DenseNet~\cite{wang2018esrgan}, without sharing parameters. 
A group of the four functions form a basic bijective ACB structure that is invertible.
Uniquely, in our design, each of the function receives the 
fusion of the image feature maps and the secret map, 
achieved by concatenating the secret map $\mathbf{K}$ 
into the image feature maps $\mathbf{X}^i$ channel-wise without losing invertibility.
That said, we replace $\omega(\mathbf{X}^i)$ in~\cite{guan2023deepmih} 
with $\omega(\mathbf{X}^i \| \mathbf{K})$ etc.
While, each nonlinear function still outputs data 
in the same dimension as the input image feature map:
\begin{equation} 
	\omega, \varphi, \rho, \eta:  \mathbb{R}^{(C + 4) \times \frac{W}{2} \times \frac{H}{2}} \mapsto \mathbb{R}^{C \times \frac{W}{2} \times \frac{H}{2}},
\end{equation}
where $C=3$ indicates the number of RGB image channels.
The transformations of each SACB are formulated as:
\begin{align}
\label{eq:forward1}
	\mathbf{Y}^{i+1} &= \mathbf{Y}^{i} \cdot \exp \left(a \left(\omega\left(\mathbf{X}^{i} \| \mathbf{K}\right)\right)\right) + \varphi\left(\mathbf{X}^{i} \| \mathbf{K}\right),\\
\label{eq:forward2}
	\mathbf{X}^{i+1} &= \mathbf{X}^{i} \cdot \exp \left(a \left(\rho\left(\mathbf{Y}^{i+1} \| \mathbf{K}\right)\right)\right) + \eta\left(\mathbf{Y}^{i+1} \| \mathbf{K}\right),
\end{align}
where $\|$ indicates channel-wise concat 
and $i \in \{1, 2, \cdots, N\}$. 
Specially, $a()$ stands for Sigmoid multiplied by a constant factor, 
which acts as activation function to constrain the processed feature map.
Being processed by $N$ sequential SACBs, the wavelet sub-bands 
of the last block are transformed back to the spatial domain to produce 
two output images:
\begin{align}
	\mathbf{\hat{x}} &= \mathrm{IWT}(\mathbf{X}^{N+1}),\\
	\mathbf{\hat{y}} &= \mathrm{IWT}(\mathbf{Y}^{N+1}),
\end{align}
where $\mathbf{\hat{y}}$ is designated as the final protection image that is 
visually identical to the pre-obfuscated image ensuring visual privacy. 
The other output $\mathbf{\hat{x}}$ is a sort of byproduct image 
which contains latent information lost during the protection process. 
Similar as~\cite{guan2023deepmih}, the byproduct $\mathbf{\hat{x}}$ is not 
required for recovery.

\subsection{Backward Recovery Flow} 
\label{sec:backward_recovery_flow}  
In the recovery pass of~\cite{guan2023deepmih}, 
a noise image sampled from Gaussian distribution is used as auxiliary 
variable to replace the latent output lost in the forward pass. 
In our framework, since random secret information following uniform distribution 
is injected and input/protection images follow the same image distribution, 
we assume the protection byproduct $\mathbf{\hat{x}}$ also approximates 
the same case-agnostic distribution as the integrated secret map $\mathbf{K}$, 
although not necessarily the identical. 
Innovatively, we opt to apply the same secret key $k$ in protection to substitute 
the lost latent output $\mathbf{\hat{x}}$ to perform image recovery. 
Therefore, the recovery process takes the protection image $\mathbf{\hat{y}}$ 
and the secret key $k$ as inputs, and generates the recovery image highly similar 
to its original. 
Instead of sampling the secret key 
at random as~\cite{guan2023deepmih}, 
we enforce the image recovery work perform correctly 
only when the correct secret key is provided, 
while making the recovery towards a wrong direction 
distinct from the original image when any different 
secret key is applied.

Illustrated in the lower part of Fig.~\ref{fig:network_design}, 
the recovery process starts with transforming the 
protected image $\mathbf{\hat{y}}$ back into the wavelet domain 
$\mathbf{\hat{Y}}^{N+1} = \mbox{DWT}(\mathbf{\hat{y}})$
and generating the secret map using the same KeyGen module 
defined in Eq.~(\ref{eq:pbkdf}) and~(\ref{eq:secret_key_map}) 
$\mathbf{K} = \mbox{KeyGen}(k)$.
The secret map $\mathbf{K}$ is repeated three times channel-wise 
to align with the dimension of the wavelet sub-bands of an RGB image 
$\mathbf{\hat{X}}^{N+1} = \mathbf{K} \| \mathbf{K} \| \mathbf{K}$,
which is treated as the auxiliary input to the backward recovery process 
to substitute the discarded protection byproduct $\mathbf{\hat{x}}$.
Then, both $\mathbf{\hat{X}}^{N+1}$ and $\mathbf{\hat{Y}}^{N+1}$ 
are processed by $N$ backward SACBs via inverted bijective functions 
based on Eq.~(\ref{eq:forward1}) and~(\ref{eq:forward2}),
with the secret map $\mathbf{K}$ integrated in the same way as protection:
\begin{align}
  \label{eq:backward1}
	\mathbf{\hat{X}}^{i} &= \left(\mathbf{\hat{X}}^{i+1} - \eta \left(\mathbf{\hat{Y}}^{i+1} \| \mathbf{K}\right)\right) 
	\cdot \exp \left(-a \left(\rho \left(\mathbf{\hat{Y}}^{i+1} \| \mathbf{K}\right)\right)\right), \\
  \label{eq:backward2}
	\mathbf{\hat{Y}}^{i} &= \left(\mathbf{\hat{Y}}^{i+1} - \varphi\left(\mathbf{\hat{X}}^{i} \| \mathbf{K}\right)\right) 
	\cdot \exp\left(-a\left(\omega\left(\mathbf{\hat{X}}^{i} \| \mathbf{K}\right)\right)\right).
\end{align}
The final output $\mathbf{\hat{X}}^{1}$ and 
$\mathbf{\hat{Y}}^{1}$ after $N$ reverse SACBs 
are transformed back to spatial domain to generate 
the recovery image $\check{\mathbf{x}}$ along with 
a recovery byproduct $\check{\mathbf{y}}$:
\begin{align}
	\check{\mathbf{x}} &= \mbox{IWT}(\mathbf{\hat{X}}^{1}),\\
	\check{\mathbf{y}} &= \mbox{IWT}(\mathbf{\hat{Y}}^{1}).
\end{align}
If any different secret key $\ddot{k}$ 
is used in the above recovery procedure, 
only a wrong recovery image $\ddot{\mathbf{x}}$ will be produced.
In both correct or wrong recovery cases, 
the recovery by-product ($\check{\mathbf{y}}$ or $\ddot{\mathbf{y}}$) 
should not disclose the original face in clear form.

\begin{figure}[t]
\centering
	\includegraphics[width=0.9\columnwidth]{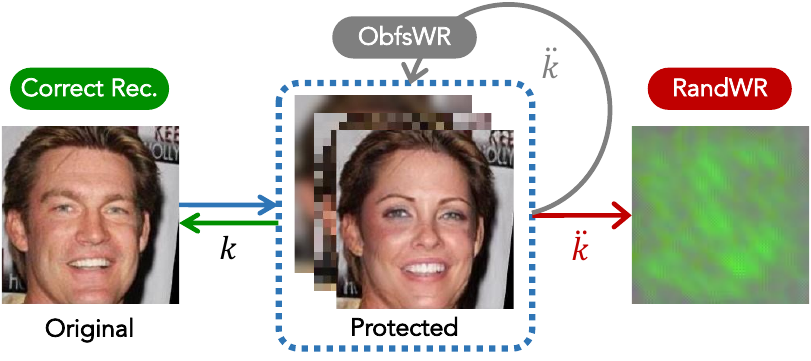}
	\caption{Illustration of the two wrong recovery modes, RandWR and ObfsWR.}
	\label{fig:recovery_modes}
\end{figure}

\subsection{Wrong Recovery Modes}  
\label{sec:recovery_modes} 
As no reference signal is available to directly 
supervise the generation of wrong recovery images,
it is not trivial to define what the wrong recovery image should look like.
Therefore, we devise two modes of wrong recovery within the 
framework to tackle malicious recovery attempts 
in different scenarios:

\subsubsection{\bf Randomized Wrong Recovery (RandWR)}
In the RandWR mode, any wrong recovery image 
presents highly randomized visual patterns, 
distinguishable from either the original or
the pre-obfuscated images significantly. 
It is hardly possible to observe any meaningful visual features 
about the original face in the wrong recovery image within this mode.
This is to simulate the incorrect decryption effect in cryptography 
where any mistake in the secret key should lead to 
totally random decryption results (aka the \textit{avalanche effect}).

\subsubsection{\bf Obfuscated Wrong Recovery (ObfsWR)} 
In the ObfsWR mode, any wrong recovery image
stays in the same visually obfuscated form as 
the protection image, or the pre-obfuscated one. 
The benefit of this mode is
making a false appearance to attackers  
that the recovery attempt has never worked, 
or the protection image is simply an obfuscated one 
without the capability of being recovered. 

The two WR modes are illustrated 
briefly in Fig.~\ref{fig:recovery_modes}. 
The design of the two  modes aims to 
offer different degrees of confusion to 
potential malicious recovery attempts. 
Exclusive training objectives are designed
to achieve the objective of each specific mode 
(in Section~\ref{sec:training_objective}). 
The invertible protection network is designed to 
work in either mode rather than supporting both at the same time. 
For a model built w.r.t. either WR mode, 
the correct recovery image using the correct secret key 
should always be near identical to the original image.

\subsection{Learning Objectives} 
\label{sec:training_objective}
To guarantee the functioning of the framework, 
a set of loss functions are designed to optimize 
the protection and recovery processes mutually with respect to 
different recovery modes. 

\subsubsection{Protection Loss} 
In the forward protection pass, the following loss is defined 
to optimize the similarity between the protection image $\mathbf{\hat{y}}$ 
and the pre-obfuscated image $\mathbf{y}$:
\begin{equation}
	\mathcal{L}_\mathrm{P} = \beta \cdot \mathrm{LPIPS}(\mathbf{\hat{y}}, \mathbf{y}) + \|\mathbf{\hat{y}} - \mathbf{y}\|_1,
\end{equation}
where $\mathrm{LPIPS}$ stands for perceptual loss defined by 
Learned Perceptual Image Patch Similarity~\cite{richard2018lpips}, 
$\|\ \|_1$ indicates L1 distance, 
and $\beta$ is a weight parameter set to 5 empirically. 

\subsubsection{Recovery Loss} 
\paragraph{Correct recovery loss}
In the recovery pass, we first apply L1 norm 
to constrain the correct recovery image in pixel level 
using the correct key applied in protection:
\begin{equation}
	\mathcal{L}_\mathrm{R} = \|\check{\mathbf{x}} - \mathbf{x}\|_1.
\end{equation}

\paragraph{RandWR loss}
Then, to optimize the functioning of the RandWR mode, 
where the wrong recovery image is expected to be randomized,
we define two special triplet losses to optimize the 
wrong recovery image simultaneously with the correct recovery image 
in a contrastive manner:
\begin{equation} \label{eq:randwr_loss}
	\mathcal{L}_\mathrm{RandWR} = \mathcal{L}_\mathrm{TriLPIPS}(\mathbf{x}, \check{\mathbf{x}},  \ddot{\mathbf{x}}) + \mathcal{L}_\mathrm{TriL1}(\mathbf{x}, \check{\mathbf{x}},  \ddot{\mathbf{x}}), 
\end{equation}
where $\mathcal{L}_\mathrm{TriLPIPS}$ stands for the triplet loss using LPIPS as distance metric 
with anchor ($A$), positive ($P$) and negative ($N$) defined as follows:
\begin{equation}
	\begin{aligned}
		\mathcal{L}_\mathrm{TriLPIPS}(&  A, P, N) =   \\ 
		 \max&\left(\mathrm{LPIPS}(A, P) - \mathrm{LPIPS}(A, N) + 1.0, 0 \right). 
	\end{aligned}
\end{equation}
Accordingly, $\mathcal{L}_\mathrm{TriL1}$ stands for similar triplet loss based on L1 norm. 
The dual triplet losses defined in Eq.~(\ref{eq:randwr_loss}) aims to push the wrong recovery image 
away from the original and the correct recovery image in both perceptual and pixel domain, 
while making the later two stay close. 
It is meaningful considering no reference signal is directly available 
for supervising the wrong recovery images in the RandWR mode.  

\paragraph{ObfsWR loss}
For the ObfsWR mode,
we use the pre-obfuscated image to supervise the 
wrong recovery image directly, 
plus dual LPIPS triplet losses 
to optimize the visual appearance of both 
correct and wrong recovery images jointly:
\begin{equation} \label{eq:obfswr_loss}
\begin{aligned}
	\mathcal{L}_\mathrm{ObfsWR} = & \\ \|\ddot{\mathbf{x}} - \mathbf{y}\|_1 + & \mathcal{L}_\mathrm{TriLPIPS}(\ddot{\mathbf{x}}, \mathbf{y}, \mathbf{x}) + \mathcal{L}_\mathrm{TriLPIPS}(\check{\mathbf{x}}, \mathbf{x}, \mathbf{y}).
	\end{aligned}
\end{equation}

\subsubsection{The overall loss}
During training, secret keys are 
sampled at random in the protection pass, 
and the same set of keys are applied in recovery 
for computing the correct recovery loss. 
In parallel, a different set of keys are 
randomly generated and used in recovery 
for computing the wrong recovery loss. 
Finally, the overall loss function for training 
the entire framework is formulated as
\begin{equation}
	\mathcal{L}_\mathrm{Total} = 
	\lambda_1 \mathcal{L}_\mathrm{P} + 
	\lambda_2 \mathcal{L}_\mathrm{R} + 
	\lambda_3 \mathcal{L}_\mathrm{WR},
\end{equation}
where $\mathcal{L}_\mathrm{WR}$ is either 
$\mathcal{L}_\mathrm{RandWR}$ or $\mathcal{L}_\mathrm{ObfsWR}$ 
depending on the wrong recovery mode expected. 
$\lambda_1$, $\lambda_2$ and $\lambda_3$ are 
hyper-parameters for balancing different loss terms.

\section{Experiments} \label{sec:experiment}


\subsection{Experimental Settings}
\subsubsection{Datasets} 
Three datasets are used in our experiments:

\textbf{CelebA}~\cite{liu2015celeba}
is one of the most widely used datasets for face recognition, 
containing 202'599 celebrity images of 10'177 identities. 
Each image is annotated with 5 landmarks and 40 binary attributes. 
The dataset is split in train, valid and test subsets.
We use its training and validation splits for training and validating our model,
and randomly sample 1'000 images from the testing split for evaluation.

\textbf{LFW}~\cite{huang2007lfw}
contains more than 13'000 images of 5'749 identities 
and provides the most widely used benchmark for face verification. 
We only use a subset of 1K images from this dataset for evaluation.

\textbf{FFHQ}~\cite{karras2019stylegan} 
contains 70'000 high-quality face images 
in resolution up to 1024$\times$1024.
The dataset exhibits diverse ages, ethnicities, 
backgrounds and accessories,
and is widely used in face generation research.
Similarly, we randomly sample 1K images from the last 
1/10 part of the dataset for testing only.

All images in our experiments are uniformly processed by 
center cropping and keeping only the facial part.
Resolution of 112$\times$112 is used throughout 
the experiments.
The sampling of the 3K test images is based on a 
fixed random seed such that the exact subset can 
be reproduced from the original datasets.
The entire 3K test set and the sampling script will 
made publicly available for future researchers 
to perform evaluations of the same task at a fair benchmark.

\def\weight{0.245}
\begin{figure*}[t]
  \subfloat{\label{sfig}\includegraphics[width=\weight\textwidth]{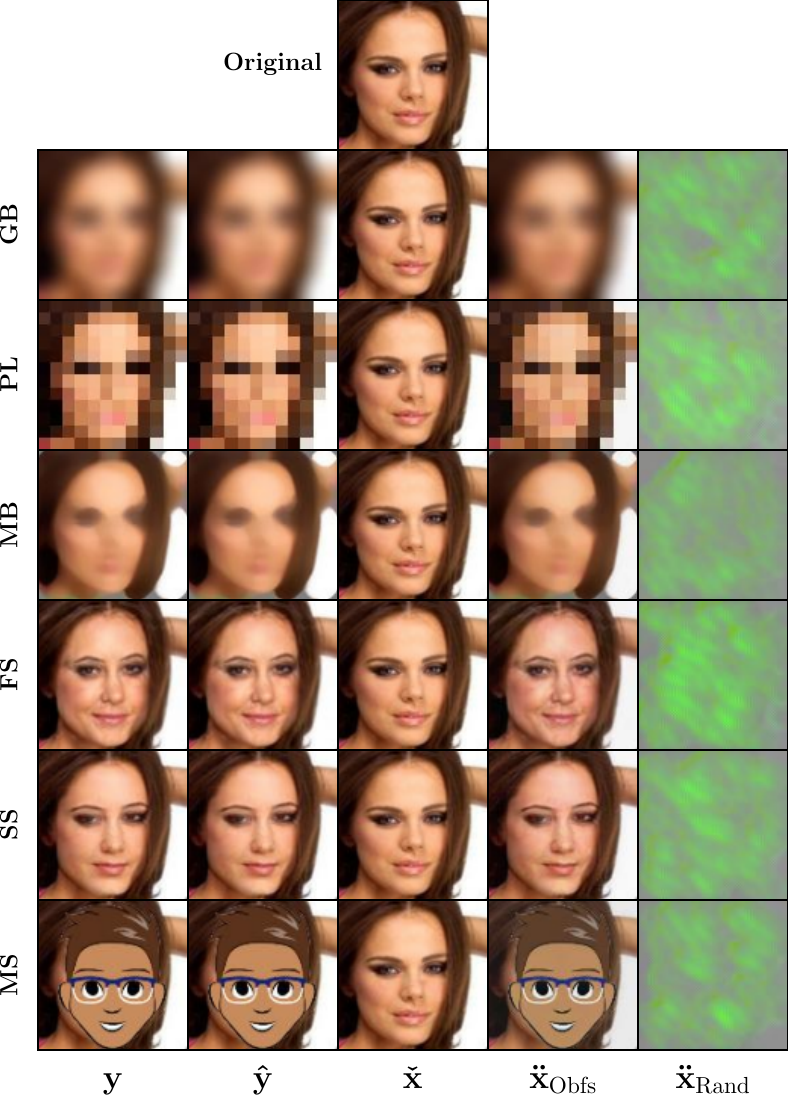}} \hfil
  \subfloat{\label{sfig}\includegraphics[width=\weight\textwidth]{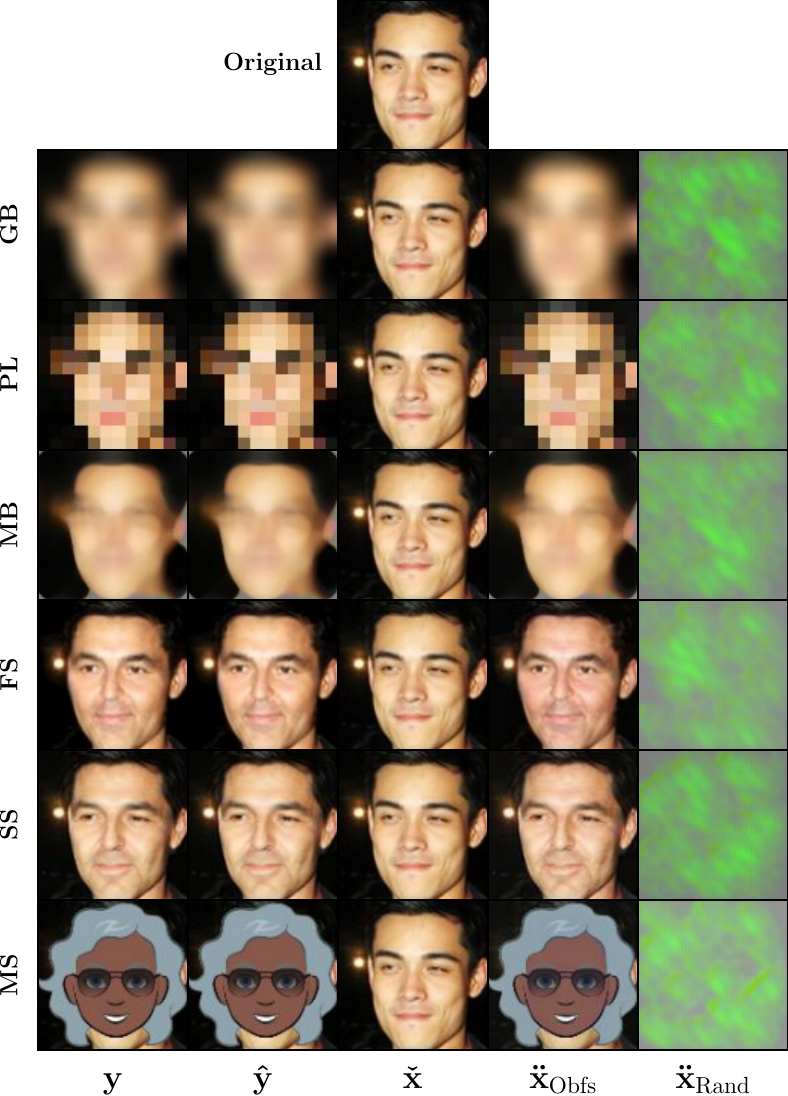}} \hfil
  \subfloat{\label{sfig}\includegraphics[width=\weight\textwidth]{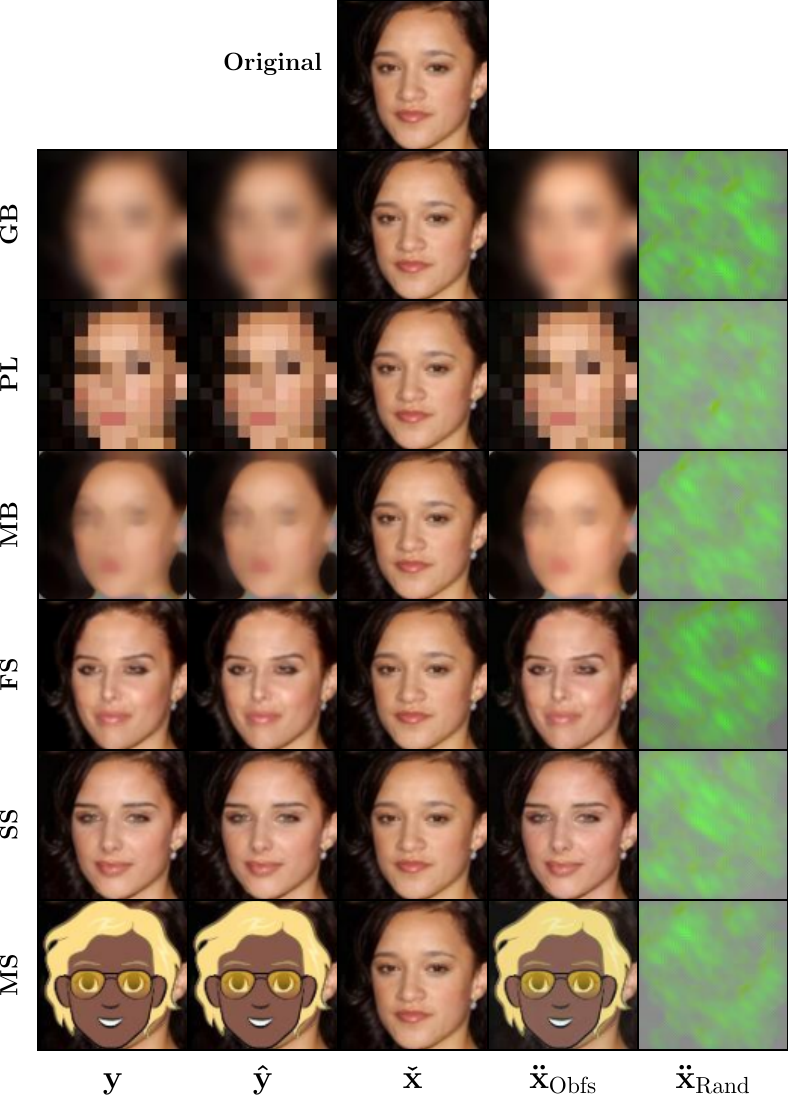}} \hfil
  \subfloat{\label{sfig}\includegraphics[width=\weight\textwidth]{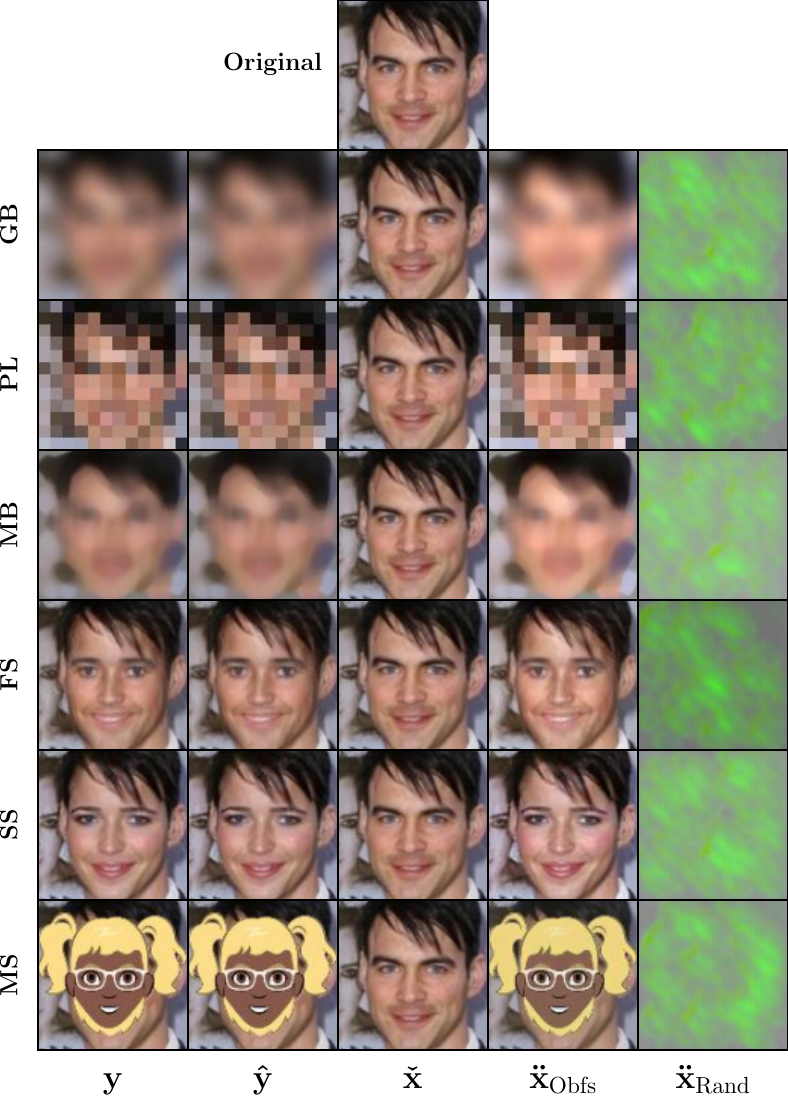}} \\
  \subfloat{\label{sfig}\includegraphics[width=\weight\textwidth]{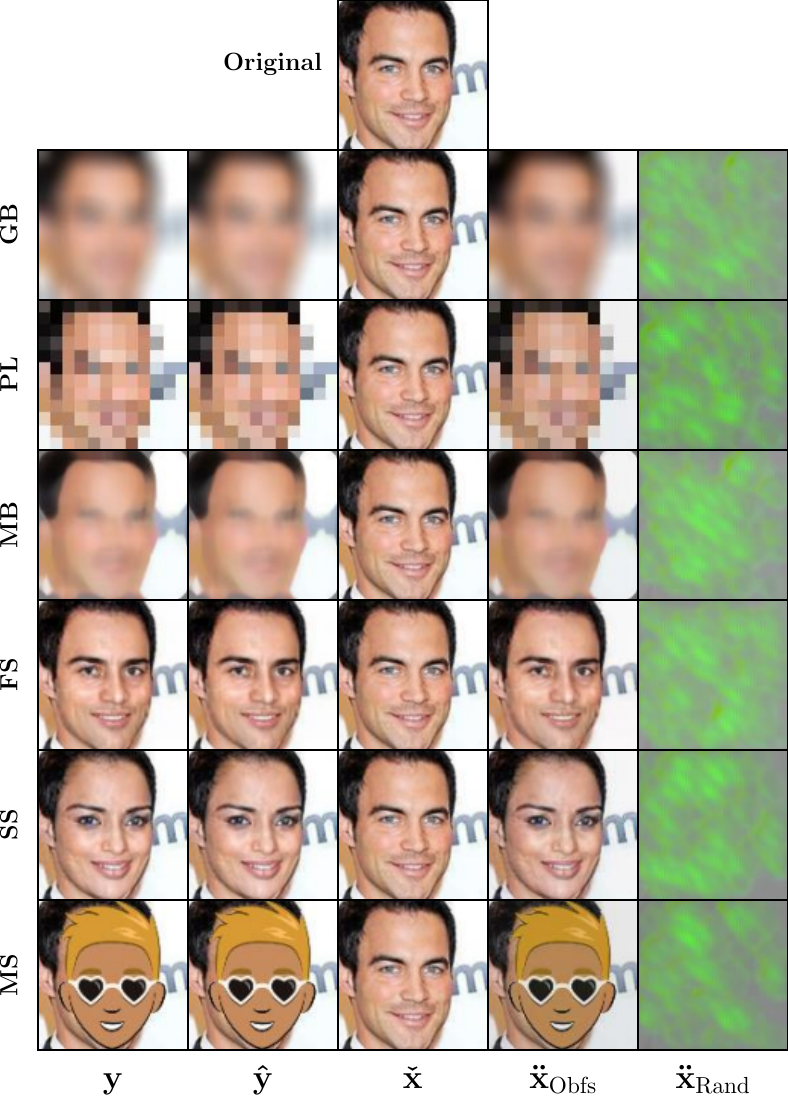}} \hfil
  \subfloat{\label{sfig}\includegraphics[width=\weight\textwidth]{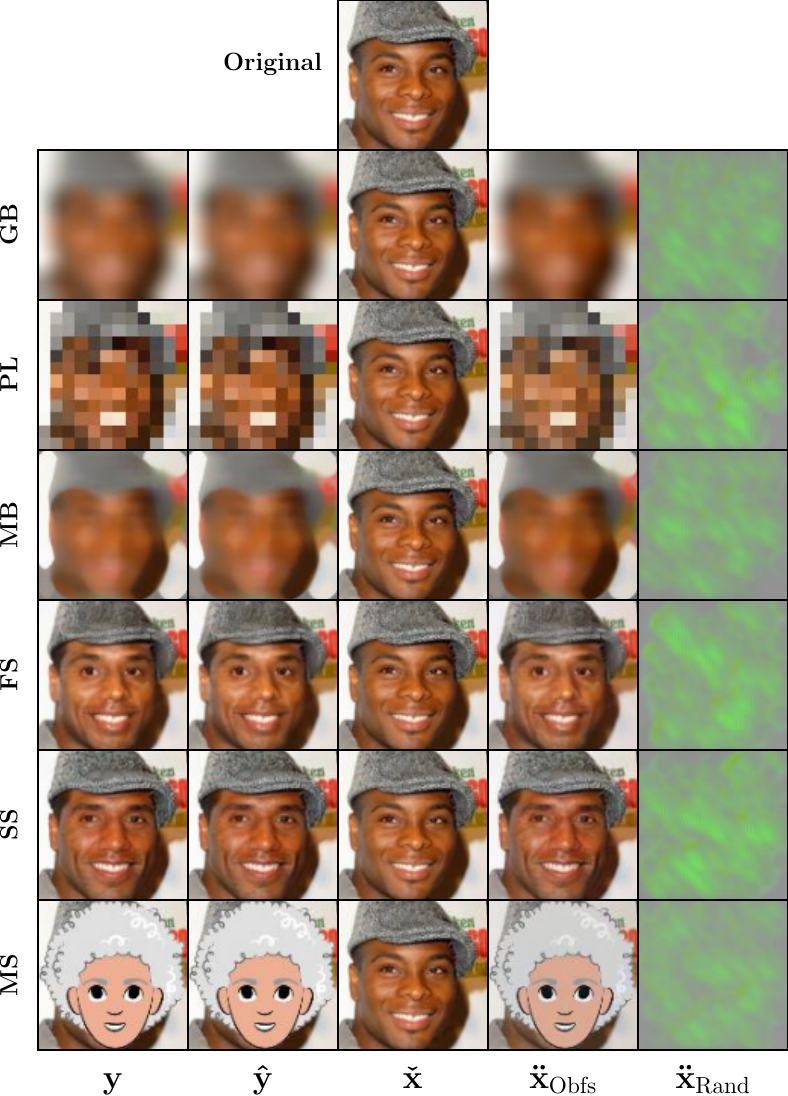}} \hfil
  \subfloat{\label{sfig}\includegraphics[width=\weight\textwidth]{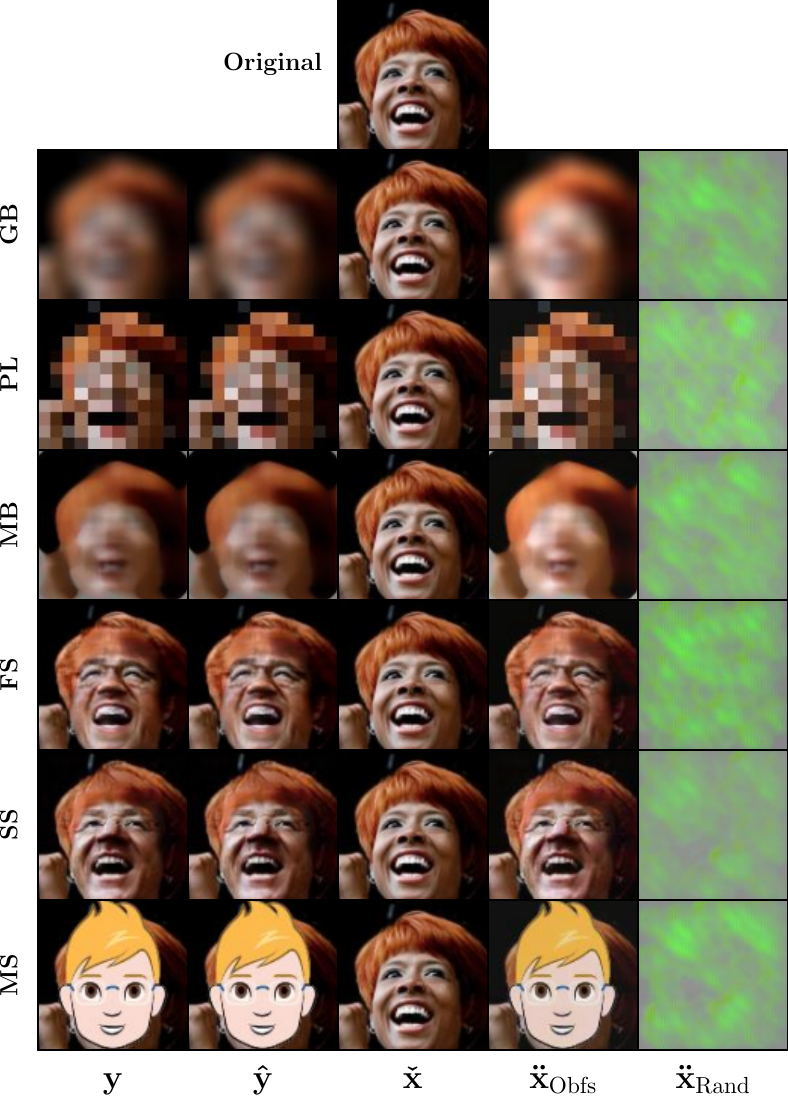}} \hfil
  \subfloat{\label{sfig}\includegraphics[width=\weight\textwidth]{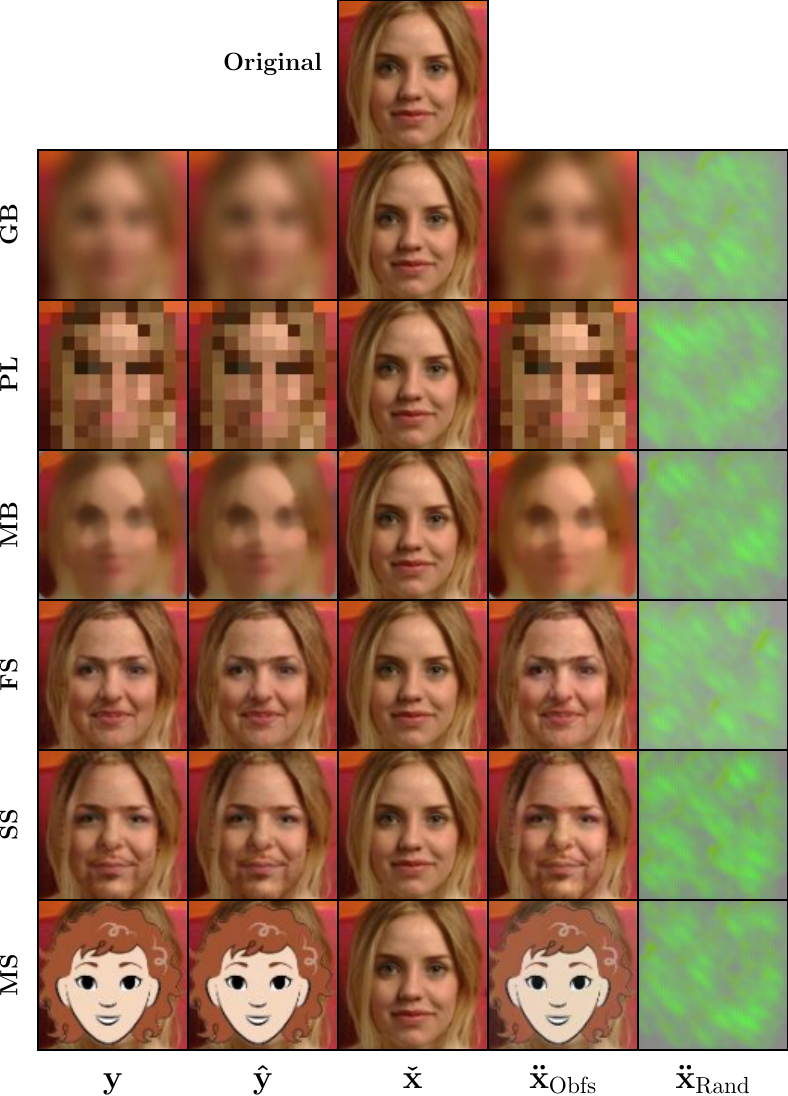}}\\
\caption{Image samples (from CelebA) within the PRO-Face S framework. On the top is the original image ($\mathbf{x}$). Columns from left to right indicate the pre-obfuscated ($\mathbf{y}$), 
protected ($\mathbf{\hat{y}}$), correct recovery ($\mathbf{\check{x}}$), and wrong recovery image in the ObfsWR ($\ddot{\mathbf{x}}_\mathrm{Obfs}$) and the RandWR ($\ddot{\mathbf{x}}_\mathrm{Rand}$) modes respectively.}
\label{fig:samples_celeba}
\end{figure*}

\subsubsection{Pre-obfuscators} 
To demonstrate the generalization of the proposed framework, 
we experiment with six visual obfuscators belonging to different types 
including Gaussian blur, pixelate, median filter, 
FaceShifter~\cite{li2020faceshifter}, 
SimSwap~\cite{chen2020simswap},
and image masking. 
The former three are typical image filters removing high-frequency visual details, 
and are by de facto the most widely used methods in practice. 
FaceShifter and SimSwap are well known face swapping algorithms 
that replace the identity of the face in image with another one.
Image masking means overlaying a cartoon facial sticker on top of the original face. 

We randomly vary the type and configuration of obfuscators 
during training and use a fixed configuration in testing. 
The specifications of the six obfuscators are given below:
\begin{itemize}
	\item {\bf Gaussian blur (GB)}: 
	the Gaussian kernel size is fixed to 21 
	while the sigma value is chosen uniformly 
	at random to lie in range of 6 to 10 during training. 
	In evaluation, the sigma value is fixed to 8.
	\item {\bf Pixelate (PL)}: the block size for pixelate is chosen 
	uniformly at random between integer 5 and 13 during training. 
	In evaluation, the block size is fixed to 9.
	\item {\bf Median blur (MB)}: the kernel size of the median filter 
	    is chosen uniformly at random between 8 and 22. 
		In evaluation, the median kernel size is fixed to 15.
	\item {\bf FaceShifter (FS)}~\cite{li2020faceshifter}: 
		the source identity images for the face swapping algorithm
		are randomly chosen from a subset of the CelebA~\cite{liu2015celeba} 
		test split that exhibit only frontal pose, in both training and testing.
	\item {\bf SimSwap (SS)}~\cite{chen2020simswap}: 
		the same as FaceShifter above.
	\item {\bf Masking (MS)}: the cartoon face sticker is randomly chosen 
	from the dataset CartoonSet~\cite{royer2020xgan} and 
	directly overlaid on the inner part of the original face, 
	in both training and testing.
\end{itemize}

\subsubsection{Metrics}
We utilize three commonly used image similarity metrics 
(PSNR, SSIM~\cite{wang2003ssim} and LPIPS~\cite{richard2018lpips})
to measure the quality of privacy protection images, 
correct recovery images, 
and the visual discrepancy of the wrong recovery images.
Unlike several prior studies where face verification 
is used to measure visual privacy, 
we keep using traditional metrics considering 
face recognition rate does not always comply with 
human subjective perception towards anonymized face, 
which has been discussed in~\cite{phillips2018pnas,yuan2022proface}. 

\paragraph{Privacy protection metric}
The visual similarity between the protection and pre-obfuscated images 
are computed to signal the \textit{\textbf{relative privacy score}}.
Note that, since our framework relies on a pre-obfuscator
as the protection template, it assumes the privacy is well preserved 
in the pre-obfuscated image.
Therefore, we call it {\it relative privacy score} as it 
indicates how the protection image is close the pre-obfuscated template, relatively. 
Higher similarity metric indicates stronger protection capability. 

\paragraph{Correct recovery metric}
The visual similarity between the correct recovery and 
the original images is computed to signal the quality of image recovery. 
Higher similarity score indicates superior recovery performance. 

\paragraph{Wrong recovery metric}
For wrong image recovery scenarios, 
we define \textit{\textbf{wrong recovery discrepancy}} of wrong recovery images 
under each WR mode as follows:
For RandWR, the discrepancy is computed as the similarity 
between the wrong recovery and the original images, 
which is expected to be low; 
For ObfsWR, the similarity metrics between the wrong recovery and 
the pre-obfuscated images are computed and expected to be high.

\subsubsection{Implementation details} 
\begin{figure}[t]
  \subfloat{\label{sfig:a}
\includegraphics[width=0.32\columnwidth]{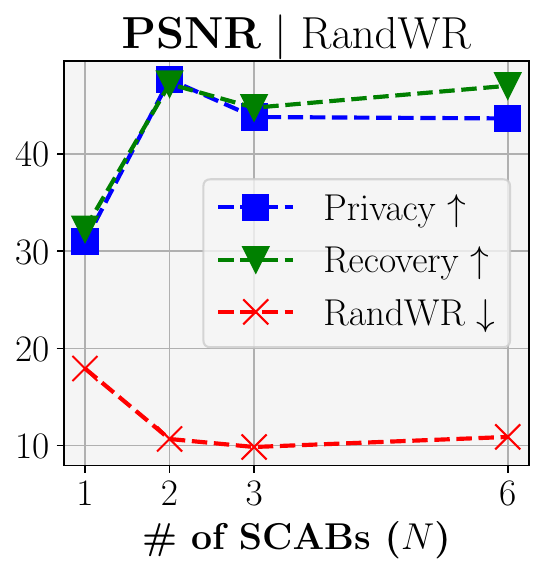}}
  \subfloat{\label{sfig:b}
\includegraphics[width=0.32\columnwidth]{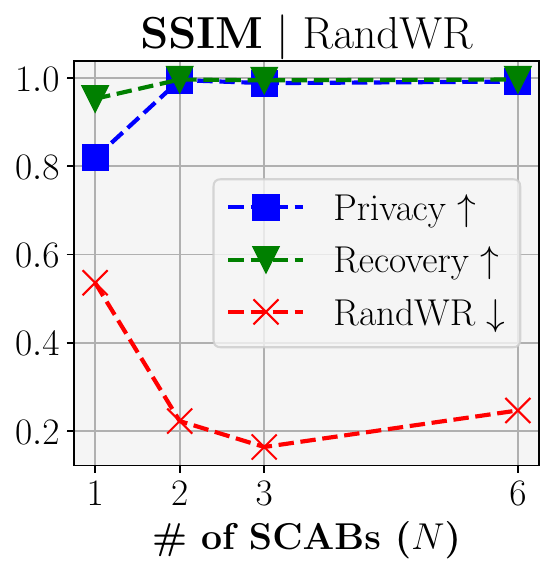}}
  \subfloat{\label{sfig:c}
\includegraphics[width=0.32\columnwidth]{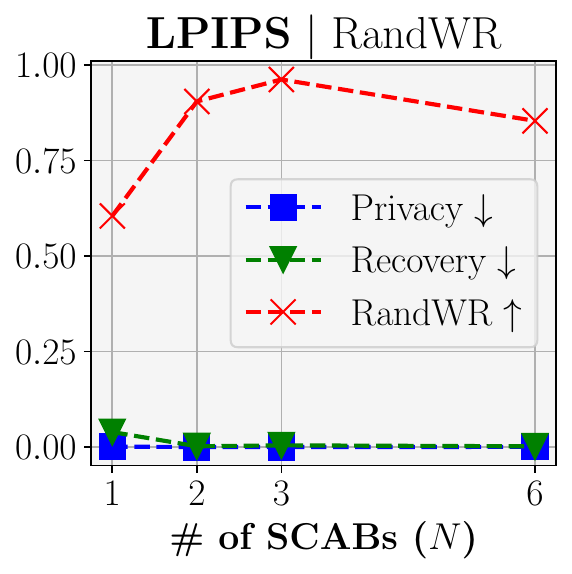}}
\\
\subfloat{\label{sfig:a}
\includegraphics[width=0.32\columnwidth]{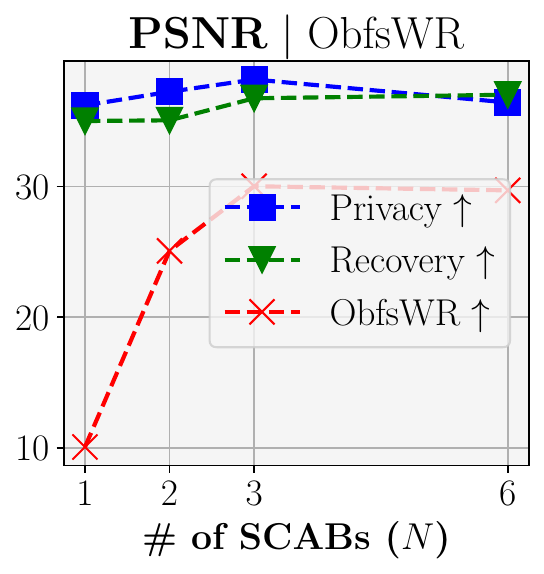}}
  \subfloat{\label{sfig:b}
\includegraphics[width=0.32\columnwidth]{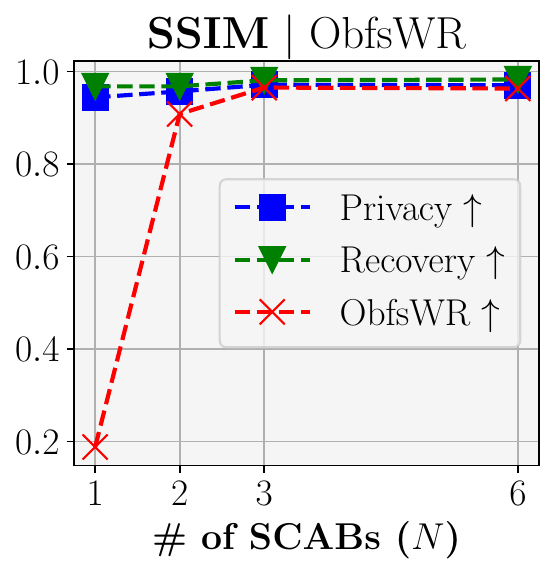}}
  \subfloat{\label{sfig:c}
\includegraphics[width=0.32\columnwidth]{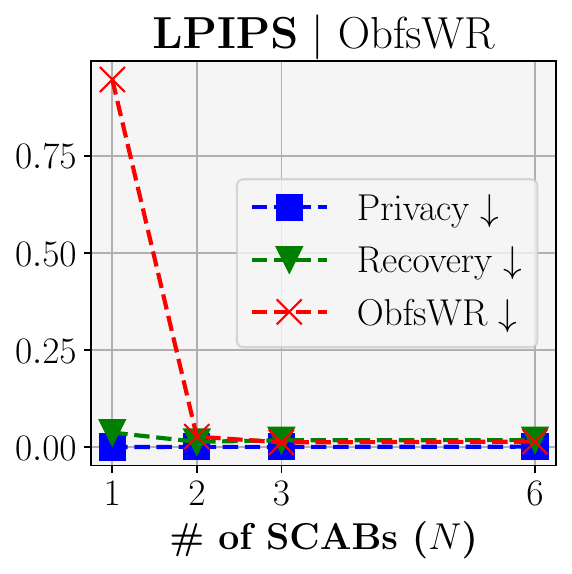}}
\caption{Protection and recovery performance on CelebA measured by PSNR, SSIM and LPIPS versus the number of SACBs.} 
\label{fig:number_of_sacbs}
\end{figure}

All experiments are performed on a single NVIDIA GeForce RTX 3090 GPU.
The Adam optimizer with $\beta_1=0.9$ and $\beta_2=0.999$ is used in training. 
A small learning rate of $0.00001$ is applied to 
avoid the vanishing gradient problem of INN. 
The training batch size is set to~12.
We experiment with different values of the hyper-parameter $N$ 
(the number of SACBs) and finally use $N$=3 as it achieves the
optimal overall performance (see Figure~\ref{fig:number_of_sacbs}) 
with relatively lightweight model size. 
The entire training process takes more than 135,000 steps 
to stabilize, which lasts approximately 12 hours. 
A visualization of several protection and recovery examples 
using different obfuscators is presented in Figure~\ref{fig:samples_celeba}.

\def\ul#1{\underline{#1}}
\def\tc{\cellcolor[HTML]{C0C0C0}}
\begin{table*}[t]
\centering
\caption{Relative privacy protection performance in terms of PSNR, SSIM and LPIPS between protection and pre-obfuscation images, measured on the test set of LFW, CelebA and FFHQ. Bold and Underlined numbers indicate the best and the second best scores over each column when comparing our averaged results (Avg.) with the other two competitors.}
\label{tab:privacy_analysis}
\begin{tabular}{llccccccccccc}
\toprule
\multicolumn{2}{c}{\multirow{2}{*}{\bf Methods}} & \multicolumn{3}{c}{\bf LFW} & \multicolumn{1}{l}{} & \multicolumn{3}{c}{\bf CelebA} & \multicolumn{1}{l}{} & \multicolumn{3}{c}{\bf FFHQ} \\ \cmidrule{3-5} \cmidrule{7-9} \cmidrule{11-13} 
\multicolumn{2}{l}{} & PSNR $\uparrow$   & SSIM $\uparrow$  & LPIPS $\downarrow$  &                      & PSNR $\uparrow$    & SSIM $\uparrow$    & LPIPS $\downarrow$  &                      & PSNR $\uparrow$   & SSIM $\uparrow$    & LPIPS $\downarrow$  \\ 
\midrule
\multirow{2}{*}{\it Competitors}
& PRO-Face~\cite{yuan2022proface}  & 34.77 & 0.973 & 0.0261 &                      & 34.95   & 0.969 & 0.0206  &                      & 34.68  & 0.966  & 0.0253  \\
& IMN~\cite{yang2023imn}  & \bf{48.79} & \ul{0.989} & 0.0079       &  & \bf{48.57} & \ul{0.988} & 0.0081       &  & \bf{48.97}    & \bf{0.991}  & 0.0068  \\
\midrule
\multirow{14}{*}{\it PRO-Face S} 
& RandWR (Avg.)        & \ul{44.92} & \bf{0.990} & \bf{0.0001}  &  & \ul{43.81} & \bf{0.989} & \bf{0.0001}  &  & \ul{43.17}    & \ul{0.988}  & \bf{0.0001}   \\
& ObfsWR (Avg.)        & 39.14      & 0.975      & \ul{0.0005}  &  & 38.20      & 0.971      & \ul{0.0006}  &  & 38.08         & 0.970       & \ul{0.0007}   \\ 
\cmidrule{2-13}
& RandWR (GB)   & 46.64 & 0.991 & 0.0    &&   45.25 & 0.987 & 0.0001   &&   45.11 & 0.986 & 0.0001 \\
& RandWR (PL)   & 44.1 & 0.992 & 0.0001    &&   43.23 & 0.99 & 0.0001   &&   42.82 & 0.989 & 0.0001 \\
& RandWR (MB)   & 44.81 & 0.991 & 0.0001    &&   43.85 & 0.988 & 0.0001   &&   43.52 & 0.987 & 0.0001 \\
& RandWR (FS)   & 47.87 & 0.996 & 0.0    &&   46.4 & 0.996 & 0.0   &&   45.04 & 0.994 & 0.0001 \\
& RandWR (SS)   & 47.97 & 0.997 & 0.0    &&   46.31 & 0.996 & 0.0   &&   44.89 & 0.995 & 0.0001 \\
& RandWR (MS)   & 38.11 & 0.975 & 0.0002    &&   37.8 & 0.973 & 0.0002   &&   37.66 & 0.973 & 0.0002 \\ 
\cmidrule{2-13}
& ObfsWR (GB)   & 42.1 & 0.974 & 0.0003    &&   41.14 & 0.968 & 0.0004   &&   41.11 & 0.967 & 0.0005 \\
& ObfsWR (PL)   & 38.92 & 0.979 & 0.0006    &&   38.15 & 0.975 & 0.0007   &&   38.13 & 0.974 & 0.0007 \\
& ObfsWR (MB)   & 40.3 & 0.976 & 0.0005    &&   39.47 & 0.97 & 0.0006   &&   39.51 & 0.969 & 0.0007 \\
& ObfsWR (FS)   & 40.99 & 0.986 & 0.0003    &&   39.72 & 0.984 & 0.0004   &&   39.42 & 0.982 & 0.0005 \\
& ObfsWR (SS)   & 40.83 & 0.986 & 0.0003    &&   39.27 & 0.985 & 0.0004   &&   39.01 & 0.983 & 0.0004 \\
& ObfsWR (MS)   & 31.69 & 0.947 & 0.001    &&   31.43 & 0.943 & 0.0011   &&   31.32 & 0.943 & 0.0012 \\ \bottomrule
\end{tabular}
\end{table*}

\subsection{Evaluation of Privacy Protection}
\begin{figure}[t]
	\includegraphics[width=1\columnwidth]{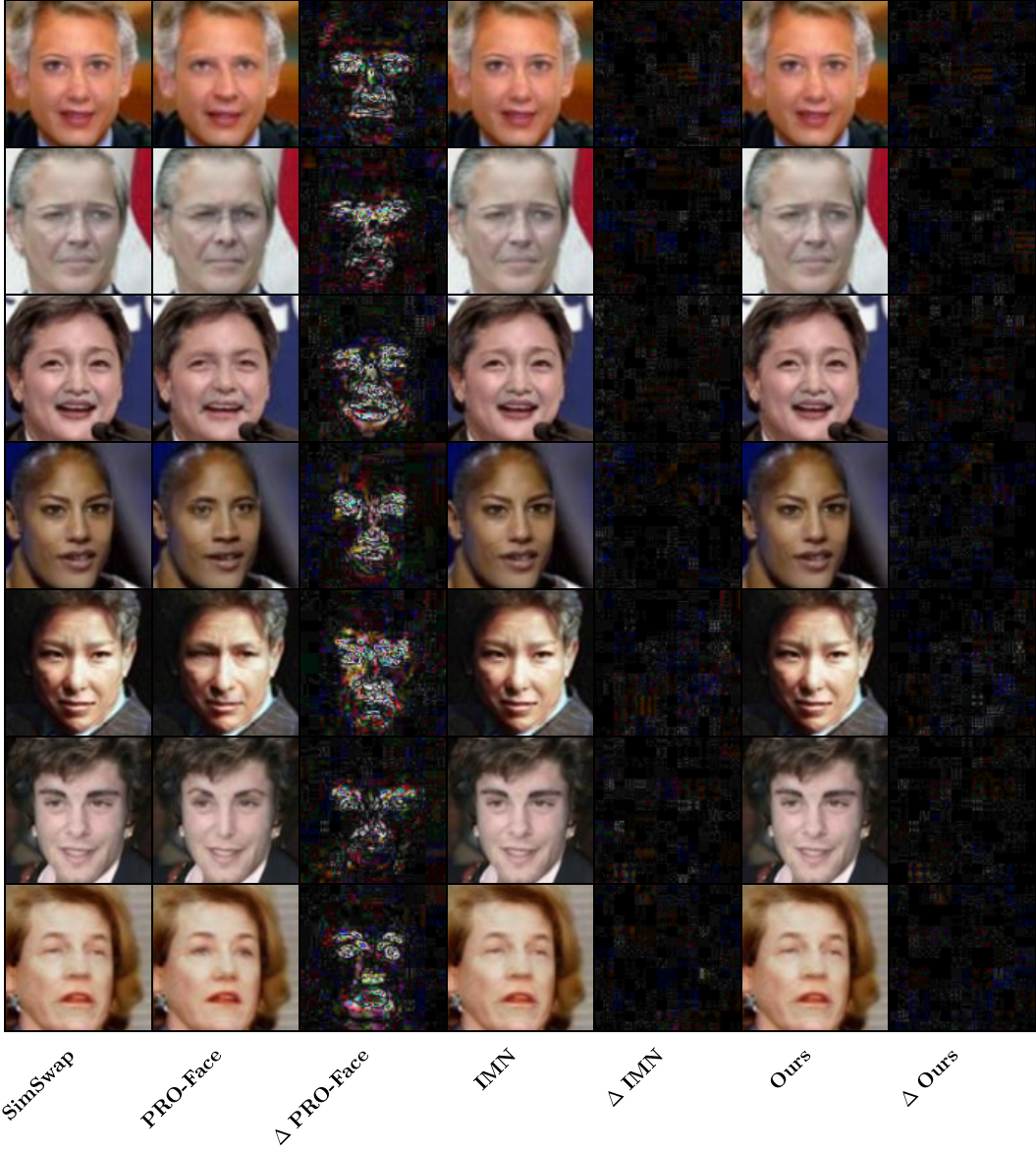}
	\caption{Protection samples (from LFW) based on SimSwap as pre-obfuscator for different methods. 
	$\Delta$ indicates $10\times$ magnified absolute difference image
	between the protection image and the SimSwap template.}
	\label{fig:plot_compare_protection_samples}
\end{figure}

To evaluate privacy, we compare with another two solutions 
employing the similar methodology of template-based protection,
i.e., PRO-Face~\cite{yuan2022proface} and IMN~\cite{yang2023imn}. 
Table~\ref{tab:privacy_analysis} shows the privacy scores of multiple cases:
for PRO-Face~\cite{yuan2022proface}, the average scores over different obfuscators are shown;
for IMN~\cite{yang2023imn}, SimSwap is used as the paper claims;
in our case, the average scores over different obfuscators and 
the scores corresponding to each obfuscater are shown. 
It is clear that our approach outperforms PRO-Face with 
large margin in most cases and metrics. 
This is because PRO-Face aims to preserve identifiable information 
in the protected images and therefore sacrifices visual privacy.
Compared with IMN~\cite{yang2023imn}, our framework built with 
the RandWR mode performs slightly better in terms of SSIM and LPIPS.
In the ObfsWR mode, our performance is slightly reduced, 
but still shows better LPIPS score than IMN~\cite{yang2023imn}. 
Nevertheless, we consider the difference between our 
approach and IMN is extremely subtle, and both should be 
considered as near lossless visual quality. 
Considering our framework is with secure reversibility 
and is much more lightweight (verified in Section~\ref{sec:complexity}), 
we claim to achieve a meaningful improvement 
over the state-of-the-art approach IMN~\cite{yang2023imn}. 
Several protection samples with SimSwap as pre-obfuscator by different methods 
are visualized in Figure~\ref{fig:plot_compare_protection_samples}.
The performance of PRO-Face~\cite{yuan2022proface} 
is visibly insufficient compared to ours. 
While, both IMN~\cite{yang2023imn} and ours are with 
excellent performance in terms of high visual similarity 
with SwimSwap obfuscation. 

When inspecting the privacy scores 
with respect to each obfuscator,
it is obvious that FaceShifter and SimSwap 
perform slightly better, 
whereas Masking ranks the last. 
This might because Cartoon sticker  
masked on the original face 
introduces much domain shift 
different from others, 
while face swapping still presents 
natural facial appearances. 
Nevertheless, such a difference in 
quality metrics is still subtle.

\subsection{Evaluation of Reversibility}
We then evaluate the reversibility of our framework, 
in terms of the correct recovery and wrong recovery performance.

\subsubsection{Correct recovery performance} 
\begin{figure}[t]
	\includegraphics[width=1\columnwidth]{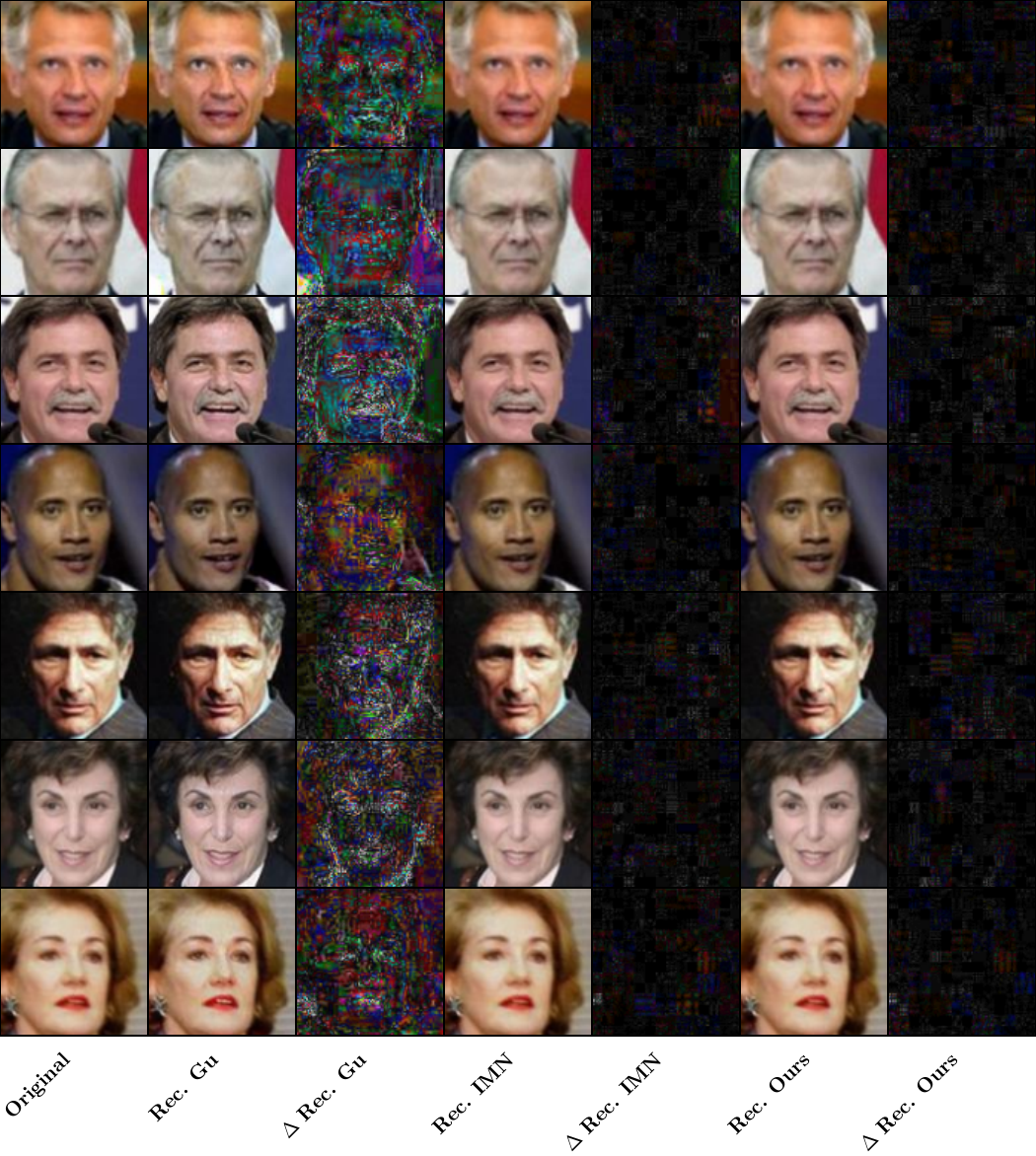}
	\caption{Recovery samples (from LFW) of different methods, where $\Delta$ Rec. indicates $10\times$ 
	magnified absolute difference image between the correct recovery image and the original image.}
	\label{fig:plot_compare_recovery_samples}
\end{figure}

In the scenario of correct image recovery, 
the recovery scores for different methods are shown in Table~\ref{tab:reovery_quality}. 
One easily observes the significant improvements of our approach over most other competitors. 
Compared with IMN~\cite{yang2023imn}, 
our approach again exhibits comparable recovery performance 
with slightly improved SSIM and LPIPS in the RandWR mode. 
The correct recovery in the mode of ObfsWR is slightly reduced 
but still keeps high quality in terms of perceptual quality 
(PSNR$>$35dB and SSIM$>$0.97).
Again, we visualize in Fig.~\ref{fig:plot_compare_recovery_samples} 
correct recovery samples of Gu~\cite{gu2020eccv}, 
IMN~\cite{yang2023imn} and ours, 
including the difference image between 
the correct recovery and the original image.
It is clear that our approach outperforms Gu~\cite{gu2020eccv} 
by presenting more precise reconstruction of color intensity and facial details, 
whereas IMN~\cite{yang2023imn} shows comparable visual quality as ours. 

\subsubsection{Wrong recovery performance}
In the scenario of wrong image recovery, 
the wrong recovery discrepancy scores for the two distinct modes respect to different 
obfuscators are listed in Table~\ref{tab:wrong_recovery_scores}.
In the RandWR mode, the discrepancy score stays in a low level as expected, 
with SSIM below 0.2 and LPIPS well above 0.9. 
In the ObfsWR mode, the discrepancy score remains high, 
indicating the success of keeping wrong recovery images obfuscated.

Once again, the similar trend is found when comparing over obfuscators, 
where Masking performs less sufficiently due to the domain difference 
induced by the cartoon sticker.

\begin{table*}[t]
\centering
\caption{Correct recovery scores in terms of PSNR, SSIM and LPIPS. Bold and Underlined numbers indicate the best and the second best scores over each column when comparing our Avg. results with the other competitors. $^\ast$We quote the recovery scores from the published version of Cao~\cite{cao2021iccv} and RAPP~\cite{zhang2023rapp} on CelebA since no open source implementation is available.}
\label{tab:reovery_quality}
\begin{tabular}{llccccccccccc}
\toprule
\multicolumn{2}{c}{\multirow{2}{*}{\bf Methods}} & \multicolumn{3}{c}{\bf LFW} & \multicolumn{1}{l}{} & \multicolumn{3}{c}{\bf CelebA} & \multicolumn{1}{l}{} & \multicolumn{3}{c}{\bf FFHQ} \\ \cmidrule{3-5} \cmidrule{7-9} \cmidrule{11-13} 
\multicolumn{2}{l}{} & PSNR $\uparrow$   & SSIM $\uparrow$  & LPIPS $\downarrow$  &                      & PSNR $\uparrow$    & SSIM $\uparrow$    & LPIPS $\downarrow$  &                      & PSNR $\uparrow$   & SSIM $\uparrow$   & LPIPS $\downarrow$  \\ 
\midrule
\multirow{5}{*}{Competitors}
& Gu~\cite{gu2020eccv}               & 30.8   & 0.938 & 0.0632 &                      & 30.52   & 0.932   & 0.0682 &                      & 28.16  & 0.921  & 0.0875 \\
& Cao~\cite{cao2021iccv}$^\ast$         & $-$     & $-$    & $-$      &                      & 27.5   & 0.902  & 0.062  &                      & $-$   & $-$  & $-$   \\
& RAPP~\cite{zhang2023rapp}$^\ast$    & $-$     & $-$    & $-$   &                      & 29.06   & 0.809    & 0.057    &                      & $-$    & $-$   & $-$   \\
& RiDDLE (e2e)~\cite{li2023riddle}    & 21.65    & 0.789   & 0.2213   &                      & 19.19   & 0.714    & 0.2285    &                      & 20.56    & 0.749   & 0.2591   \\
& RiDDLE (latent)~\cite{li2023riddle} & 27.14  & 0.921 & 0.0941  &                      & 26.2    & 0.9   & 0.089  &                      & 27.14   & 0.921  & 0.0941 \\ 
& IMN~\cite{yang2023imn}              & \bf{49.1}  & \ul{0.995} & 0.0210       && \bf{49.14} & \ul{0.995} & \ul{0.0175}  && \bf{48.83}  & \bf{0.995}  & \ul{0.0172}  \\
\midrule
\multirow{14}{*}{PRO-Face S}
& RandWR (Avg.) & \ul{46.87} & \bf{0.998} & \bf{0.0044}  && \ul{44.76} & \bf{0.997} & \bf{0.0043}  && \ul{41.14}  & \ul{0.990}  & \bf{0.0061}  \\
& ObfsWR (Avg.) & 37.95      & 0.985      & \ul{0.0170}  && 36.76      & 0.981      & 0.0176       && 35.86       & 0.977      & 0.0194       \\ 
\cmidrule{2-13} 
& RandWR (GB)   & 49.18 & 0.998 & 0.0009    &&   46.37 & 0.997 & 0.0014   &&   42.03 & 0.992 & 0.0031 \\
& RandWR (PL)   & 46.91 & 0.997 & 0.0021    &&   44.9 & 0.996 & 0.0024   &&   41.19 & 0.991 & 0.0044 \\
& RandWR (MB)   & 44.36 & 0.998 & 0.0079    &&   42.98 & 0.996 & 0.0067   &&   40.3 & 0.992 & 0.0073 \\
& RandWR (FS)   & 49.49 & 0.998 & 0.0009    &&   46.77 & 0.997 & 0.0013   &&   42.25 & 0.993 & 0.0031 \\
& RandWR (SS)   & 49.05 & 0.998 & 0.001    &&   46.27 & 0.997 & 0.0015   &&   41.87 & 0.992 & 0.0034 \\
& RandWR (MS)   & 42.2 & 0.987 & 0.0139    &&   41.26 & 0.986 & 0.0127   &&   39.22 & 0.981 & 0.0153 \\
\cmidrule{2-13}
& ObfsWR (GB)   & 37.7 & 0.984 & 0.0167    &&   36.06 & 0.977 & 0.0191   &&   35.3 & 0.975 & 0.022 \\
& ObfsWR (PL)   & 38.82 & 0.985 & 0.0149    &&   37.52 & 0.982 & 0.016   &&   36.43 & 0.978 & 0.0178 \\
& ObfsWR (MB)   & 37.5 & 0.984 & 0.0191    &&   35.95 & 0.978 & 0.0215   &&   35.18 & 0.975 & 0.0238 \\
& ObfsWR (FS)   & 38.13 & 0.986 & 0.017    &&   37.33 & 0.984 & 0.0163   &&   36.31 & 0.98 & 0.0166 \\
& ObfsWR (SS)   & 38.13 & 0.988 & 0.018    &&   37.08 & 0.985 & 0.0177   &&   36.12 & 0.981 & 0.0196 \\
& ObfsWR (MS)   & 37.41 & 0.98 & 0.0159    &&   36.64 & 0.978 & 0.0152   &&   35.8 & 0.975 & 0.0165 \\
\bottomrule
\end{tabular}
\end{table*}

\begin{table*}[t]
\centering
\caption{Wrong recovery discrepancy in terms of PSNR, SSIM and LPIPS with respect to different WR modes and obfuscators.}
\label{tab:wrong_recovery_scores}
\begin{tabular}{ccccccccccccc}
\toprule
\multicolumn{2}{c}{\bf WR mode / Obf.} & \multicolumn{3}{c}{\bf LFW} &  & \multicolumn{3}{c}{\bf CelebA} &  & \multicolumn{3}{c}{\bf FFHQ} \\ 
\midrule                    
\multicolumn{2}{l}{}  & PSNR $\downarrow$   & SSIM $\downarrow$  & LPIPS $\uparrow$  &                      & PSNR $\downarrow$   & SSIM $\downarrow$  & LPIPS $\uparrow$  &                      & PSNR $\downarrow$   & SSIM $\downarrow$  & LPIPS $\uparrow$   \\ 
\cmidrule{3-5} \cmidrule{7-9} \cmidrule{11-13}
\multirow{6}{*}{RandWR} 
& Avg. & 10.45 & 0.183 & 0.9454    &&   9.86 & 0.164 & 0.9616   &&   10.53 & 0.186 & 0.9988 \\
& GB   & 10.44 & 0.179 & 0.9494    &&   9.86 & 0.16  & 0.9680   &&   10.54 & 0.183 & 1.0037 \\
& PL   & 10.46 & 0.183 & 0.9454    &&   9.89 & 0.165 & 0.9603   &&   10.53 & 0.184 & 0.9997 \\
& MB   & 10.45 & 0.179 & 0.9489    &&   9.86 & 0.161 & 0.9650   &&   10.51 & 0.183 & 1.0027 \\
& FS   & 10.47 & 0.182 & 0.9465    &&   9.87 & 0.163 & 0.9617   &&   10.56 & 0.186 & 0.9993 \\
& SS   & 10.47 & 0.182 & 0.9467    &&   9.86 & 0.163 & 0.9628   &&   10.53 & 0.184 & 1.0011 \\
& MS   & 10.39 & 0.192 & 0.9355    &&   9.8 & 0.171 & 0.952   &&   10.54 & 0.198 & 0.9864 \\
\midrule
\multicolumn{2}{l}{}  & PSNR $\uparrow$   & SSIM $\uparrow$  & LPIPS $\downarrow$  &                      & PSNR $\uparrow$    & SSIM $\uparrow$    & LPIPS $\downarrow$  &                      & PSNR $\uparrow$   & SSIM $\uparrow$    & LPIPS $\downarrow$   \\ 
\cmidrule{3-5} \cmidrule{7-9} \cmidrule{11-13} 
\multirow{6}{*}{ObfsWR}
& Avg. & 28.97 & 0.968 & 0.0147    &&   30.03 & 0.965 & 0.0122   &&   28.76 & 0.971 & 0.0153 \\
& GB   & 24.57 & 0.966 & 0.0281    &&   27.03 & 0.969 & 0.0212   &&   25.49 & 0.969 & 0.0291 \\
& PL   & 28.57 & 0.963 & 0.0102    &&   29.73 & 0.959 & 0.0085   &&   28.36 & 0.969 & 0.0103 \\
& MB   & 25.39 & 0.96  & 0.021     &&   27.65 & 0.960 & 0.0168   &&   25.9  & 0.965 & 0.0212 \\
& FS   & 33.64 & 0.977 & 0.0095    &&   34.10 & 0.970 & 0.0082   &&   32.48 & 0.982 & 0.0104 \\
& SS   & 34.55 & 0.985 & 0.0097    &&   34.82 & 0.982 & 0.0088   &&   33.3  & 0.986 & 0.0094 \\
& MS   & 27.12 & 0.955 & 0.0098    &&   26.83 & 0.949 & 0.01     &&   27.02 & 0.957 & 0.0115 \\
\bottomrule
\end{tabular}
\end{table*}


\begin{figure*}[t]
  \subfloat{\label{sfig:security_analysis_wrong_recovery}
\includegraphics[height=0.44\textwidth]{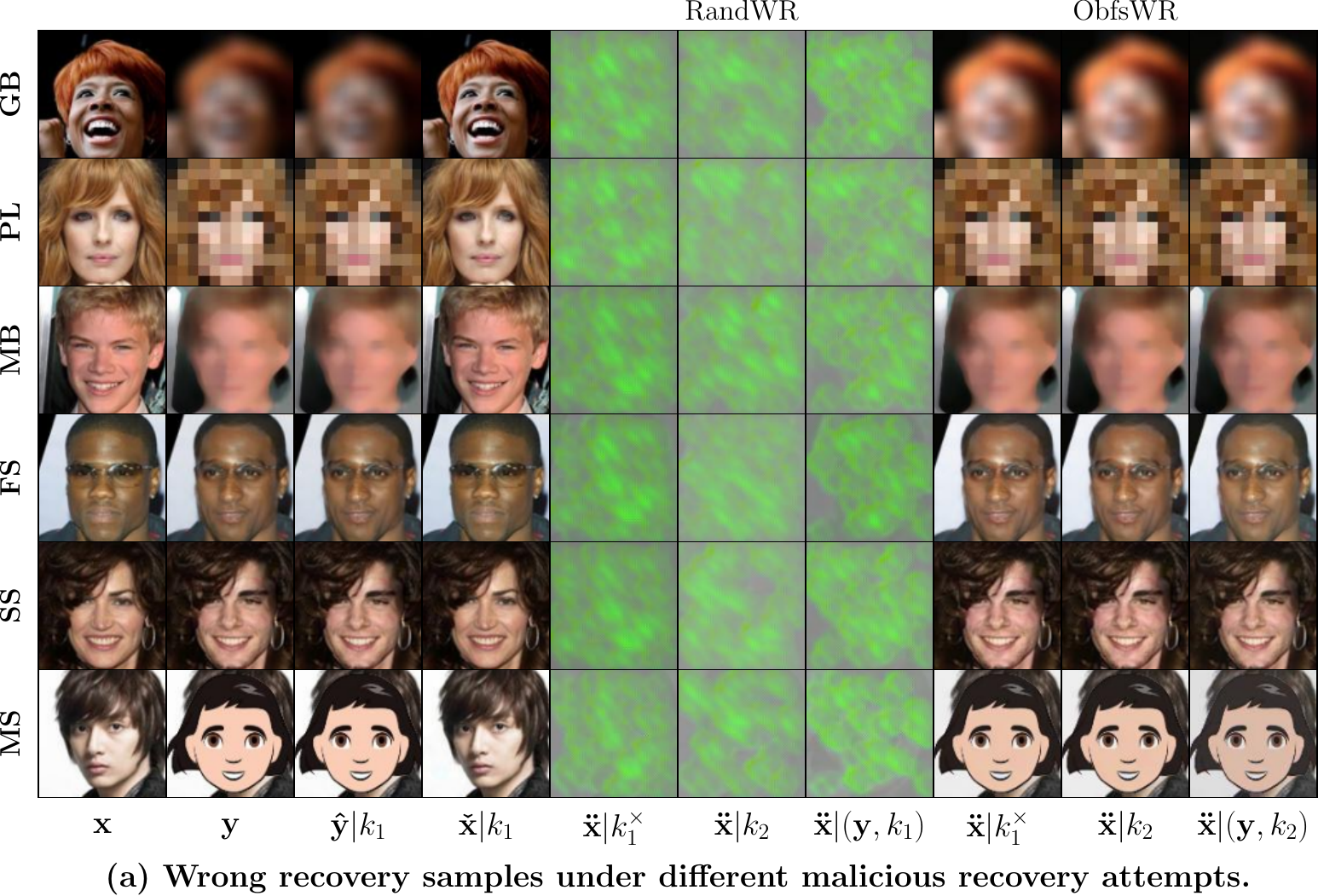}}
\hfil
  \subfloat{\label{sfig:security_analysis_byproducts}
\includegraphics[height=0.44\textwidth]{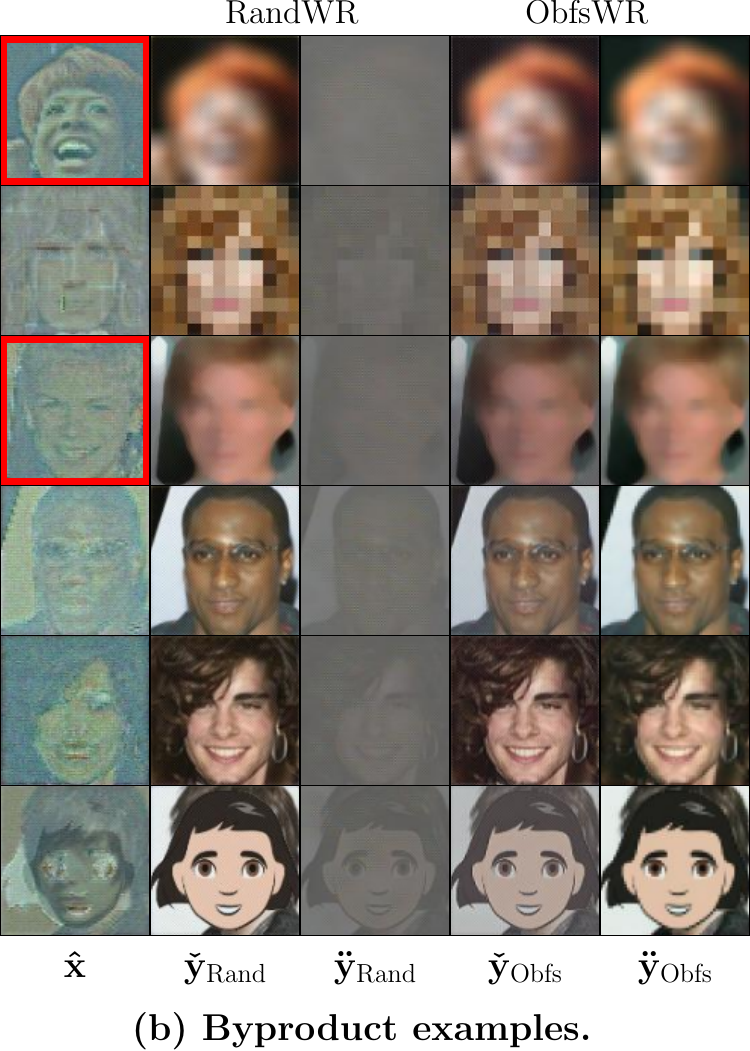}}
\caption{Image samples (from FFHQ) illustrating (a) wrong recovery under various malicious recovery attempts, 
and (b) byproducts generated within the framework. 
$k_1^\times$ denotes a secret key that has only 1-bit difference from $k_1$, 
and $k_2$ denotes a random secret key significantly different from $k_1$.}
\label{fig:security_analysis}
\end{figure*}


\subsection{Security Analysis}
Since PRO-Face~S does not claim to 
offer rigorous security as cryptography,
we analyze the security of the proposed 
framework in an empirical and qualitative 
way from the following two aspects: 

\subsubsection{Malicious recovery attempts} 
We first experiment with ``malicious'' 
recovery attempts using wrong secret keys or 
by substituting the protection image with the pre-obfuscated one 
to simulate a kind of ``spoofing attack''. 
Several wrong recovery samples under the two cases are shown in 
Fig.~\ref{sfig:security_analysis_wrong_recovery}.
It is obvious that the recovery still fails even with 1-bit difference 
in the secret key, for both RandWR and ObfsWR cases.
When presenting the pre-obfuscated image to the recovery module, 
it is still hard to recover the correct image although 
the pre-obfuscated image is highly similar to the protection image. 
In practice, this feature allows us to apply the full protection procedure only 
on a few key video frames while keeping the other frames in pre-obfuscated form, 
similar as PECAM~\cite{wu2021pecam}.
It will not only reduce computational burdens 
but also confuse potential attackers.

\subsubsection{Byproduct analysis} \label{sec:byproduct_analysis}
We also analyze the security risks potentially existing in the 
byproduct images generated along protection and recovery, 
which has never been thoroughly investigated in prior studies 
employing the similar methodologies, e.g., IMN~\cite{yang2023imn}. 
Hereby, we visualize in Fig.~\ref{sfig:security_analysis_byproducts} 
several samples of the protection byproduct $\hat{\mathbf{x}}$, and
the recovery byproduct in both correct and wrong recovery cases, i.e.,  
$\check{\mathbf{y}}_\mathrm{Rand/Obfs}$ and $\ddot{\mathbf{y}}_\mathrm{Rand/Obfs}$.
It is clear that any byproduct image generated in the recovery
stage does not disclose the original face in clear form. 
However, we do recognize the risk existing in the byproduct 
$\hat{\mathbf{x}}$ generated along protection, 
which reveals certain level visual information 
about the original face, especially with the blur-based 
obfuscators (marked in red rectangle). 
Considering the protection byproduct output is not required for recovery, 
it should be discarded immediately upon generation 
to minimize the privacy risks.

\subsection{Complexity and Cost} \label{sec:complexity}
To demonstrate the lightweight of our framework, 
we summarize the number of model parameters and 
the inference time for multiple reversible 
solutions in Table~\ref{tab:storage_and_cost}. 
Image size and batch size are uniformly set to 112$\times$112 
and 16 for a fair comparison. 
For PRO-Face~S and IMN~\cite{yang2023imn}, 
we count only the model size of the 
invertible network without considering the pre-obfuscation model.
It is clear that PRO-Face~S possesses the minimum 
number of model parameters ($\sim$1M) among the all. 
As for the time cost, it takes only 0.014 seconds per batch for our 
framework to finish a full protection operation using Gaussian blur as pre-obfuscator. 
When using the more sophisticated SimSwap, 
the inference time is increased to 0.595 seconds but still 
well below the other competitors. 
Therefore, we claim the proposed framework is lightweight 
yet offering the optimal overall performance.

\begin{table*}[t]
\centering
\caption{Results of the ablation study on privacy protection, correct and wrong recoveries with respect to different framework variations. The numbers in bold indicate the best score over each column.}
\label{tab:ablation_study}
\setlength{\tabcolsep}{4.5pt}
\begin{tabular}{lccccccccccccccc}
\toprule
\multirow{2}{*}{}  & \multicolumn{3}{c}{\bf Privacy protection} &  & \multicolumn{3}{c}{\bf Correct recovery} &  & \multicolumn{3}{c}{\bf RandWR} &  & \multicolumn{3}{c}{\bf ObfsWR} \\ \cmidrule{2-4} \cmidrule{6-8} \cmidrule{10-12} \cmidrule{14-16} 
                   & PSNR $\uparrow$      & SSIM $\uparrow$       & LPIPS $\downarrow$      &  & PSNR $\uparrow$      & SSIM $\uparrow$      & LPIPS $\downarrow$      &  & PSNR $\downarrow$   & SSIM $\downarrow$    & LPIPS $\uparrow$  &  & PSNR $\uparrow$  & SSIM $\uparrow$   & LPIPS $\downarrow$   \\ 
\midrule
w/o $\mathcal{L}_\mathrm{TriL1}$          & 41.70      & 0.985       & 0.0004      &  & 43.34      & 0.992     & 0.0045     &  & 10.68  &  0.284  & 0.7866  &  & $-$   & $-$   & $-$   \\ 
w/o $\mathcal{L}_\mathrm{TriLPIPS}$       & 39.76      & 0.934       & 0.0021       &  & 41.01      & 0.946      & 0.0137      &  & $-$   & $-$   & $-$   &  & 28.61  & 0.923   & 0.0201   \\ 
w/o $\mathbf{K}$ in rec.      & 42.36      & 0.984       & 0.0001       &  & 42.03      & 0.990      & 0.0084     &  & 10.12   & 0.252   & 0.828   &  & 29.68   & 0.961  & 0.0131  \\ 
Full               & {\bf 43.81}      & {\bf 0.989}     & {\bf 0.0001}     &  & {\bf 44.76}     & {\bf 0.995}    & {\bf 0.0043}    &  & {\bf 9.86} &  {\bf 0.164}   & {\bf 0.9616}   &  & {\bf 30.03}  & {\bf 0.965}  & {\bf 0.0122}  \\ 
\bottomrule
\end{tabular}
\end{table*}

\subsection{Ablation Study}
Last but not least, we carry our the ablation study 
to verify the effectiveness of several key components 
of the proposed framework.
Instead of checking the effectiveness of 
the DWT/IWT modules and the DenseNet structure in the INN 
that have been well investigated in~\cite{guan2023deepmih,yang2023imn}, 
we analyze the functionality of several unique components introduced in
the PRO-Face S: 
The first variation is the triplet loss $\mathcal{L}_\mathrm{TriL1}$ defined 
in Eq.~(\ref{eq:randwr_loss}) for RandWR mode optimization; 
The second are the two perceptual triplet losses $\mathcal{L}_\mathrm{TriLPIPS}$ 
specified in Eq.~(\ref{eq:obfswr_loss}) for the ObfsWR mode; 
Another variation is the use of the secret map $\mathbf{K}$ 
as input of recovery instead of using Gaussian noise as~\cite{guan2023deepmih,yang2023imn}. 
We train and test the framework by removing each of those variations and report  
the performance on CelebA in Table~\ref{tab:ablation_study}.
It is quite clear that our final configuration integrating 
all above variations offers the optimal overall performance,
which verifies the effectiveness of each introduced component.

\section{Conclusion} \label{sec:conclusion}
This paper proposes {\bf PRO-Face S}, a unified framework for 
\textbf{P}rivacy-preserving \textbf{R}eversible 
\textbf{O}bfuscation of \textbf{Face} images via 
\textbf{S}ecure flow-based model. 
The framework elegantly assembles multiple merits 
that a visual privacy protection solution should have, 
including higher diversity supporting different types of obfuscation, 
personalized anonymity powered by use-specified obfuscator, 
near lossless image recovery thanks to INN, 
security against different malicious recovery attempts, 
and yet lightweight model allowing for more practical applications. 
Extensive experiments and comprehensive analysis 
demonstrate the effectiveness of the proposed framework 
in privacy protection, image recovery and empirical security enforcement.

\begin{table}[t]
\centering
\caption{Storage and time cost of different reversible solutions.} 
\label{tab:storage_and_cost}
\setlength{\tabcolsep}{4pt}
\begin{tabular}{ccccc}
\toprule
{\bf Metric}     & {\bf PRO-Face S}      & {\bf IMN}~\cite{yang2023imn}   & {\bf Gu}~\cite{gu2020eccv}     & {\bf Riddle}~\cite{li2023riddle} \\ 
\midrule
\# params  & 1.07M           & 4.05M & 11.43M & 40.03M       \\ 
s/batch &  min 0.014 / max 0.595  & 1.84  & 1.229  & 2.49    \\ 
\bottomrule
\end{tabular}
\end{table}

Yet, certain limitations still remain, which have not been addressed in our work.
For instance, we make a strong assumption that the pre-obfuscated image 
is privacy-free, which is not always true as the obfuscated image 
may still be inverted to clear form by image enhancement.
The recovery robustness against commonly used image processing 
(such as JPEG) applied on the protection image is still an open question. 
Moreover, it is ideal to solve the privacy risk related to the protection byproduct 
mentioned in Section~\ref{sec:byproduct_analysis} to build a more secure framework.
Those will serve as our future work.
%
%
%
%
%
%
%
%

\bibliographystyle{IEEEtran}
\bibliography{mm2023}

\begin{IEEEbiography}[{\includegraphics[width=1in,height=1.25in,clip,keepaspectratio]{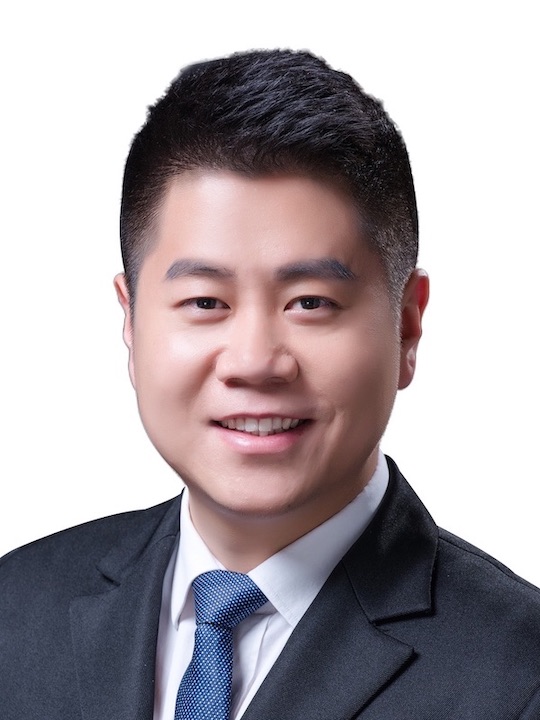}}]{Lin Yuan}
received the B.Eng. degree in electronic science and tchnology from the University of Electronic Science and Technology of China (UESTC) in 2011 and the Ph.D. degree in electrical engineering from École Polytechnique Fédérale de Lausanne (EPFL), Switzerland in 2017. 
He is currently working as researcher at the School of Cyber Security and Information Law in Chongging University of Posts and Telecommunications. 
His research interests include image and video analysis, multimedia privacy protection, and media forensics.
\end{IEEEbiography}


\begin{IEEEbiography}
[{\includegraphics[width=1in,height=1.25in,clip,keepaspectratio]{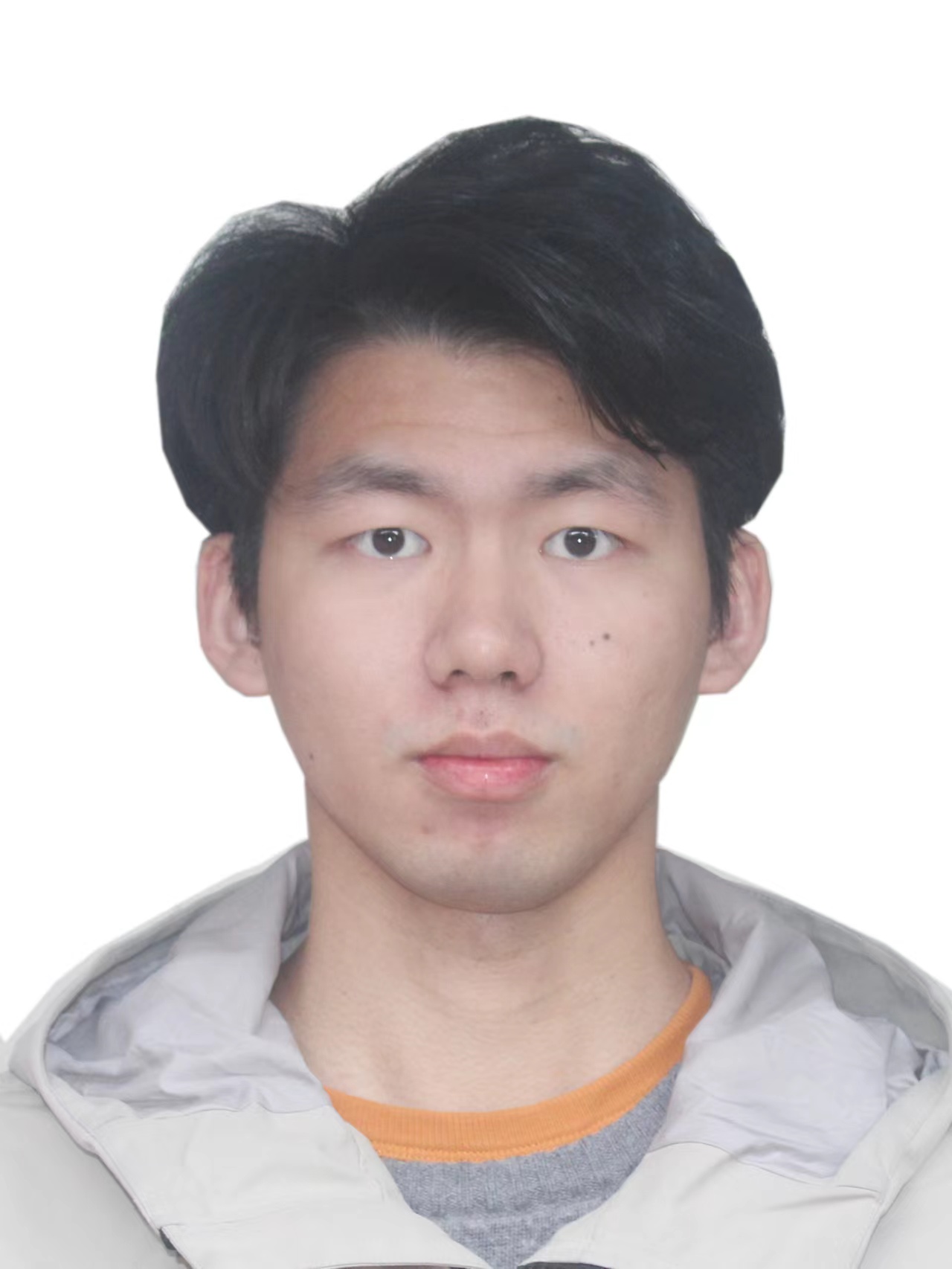}}]{Kai Liang}
received the B.S. degree in electronic information engineering from the School of Electrical and Electronic Engineering, Chongqing University of Technology, Chongqing, China, in June 2021. He is currently pursuing the M.S. degree in electronic science and technology from the School of Photoelectric Engineering, Chongqing University of Posts and Telecommunications, Chongqing, China. His research interests include face privacy  protection and face editing. 
\end{IEEEbiography}


\begin{IEEEbiography}[{\includegraphics[width=1in,height=1.25in,clip,keepaspectratio]{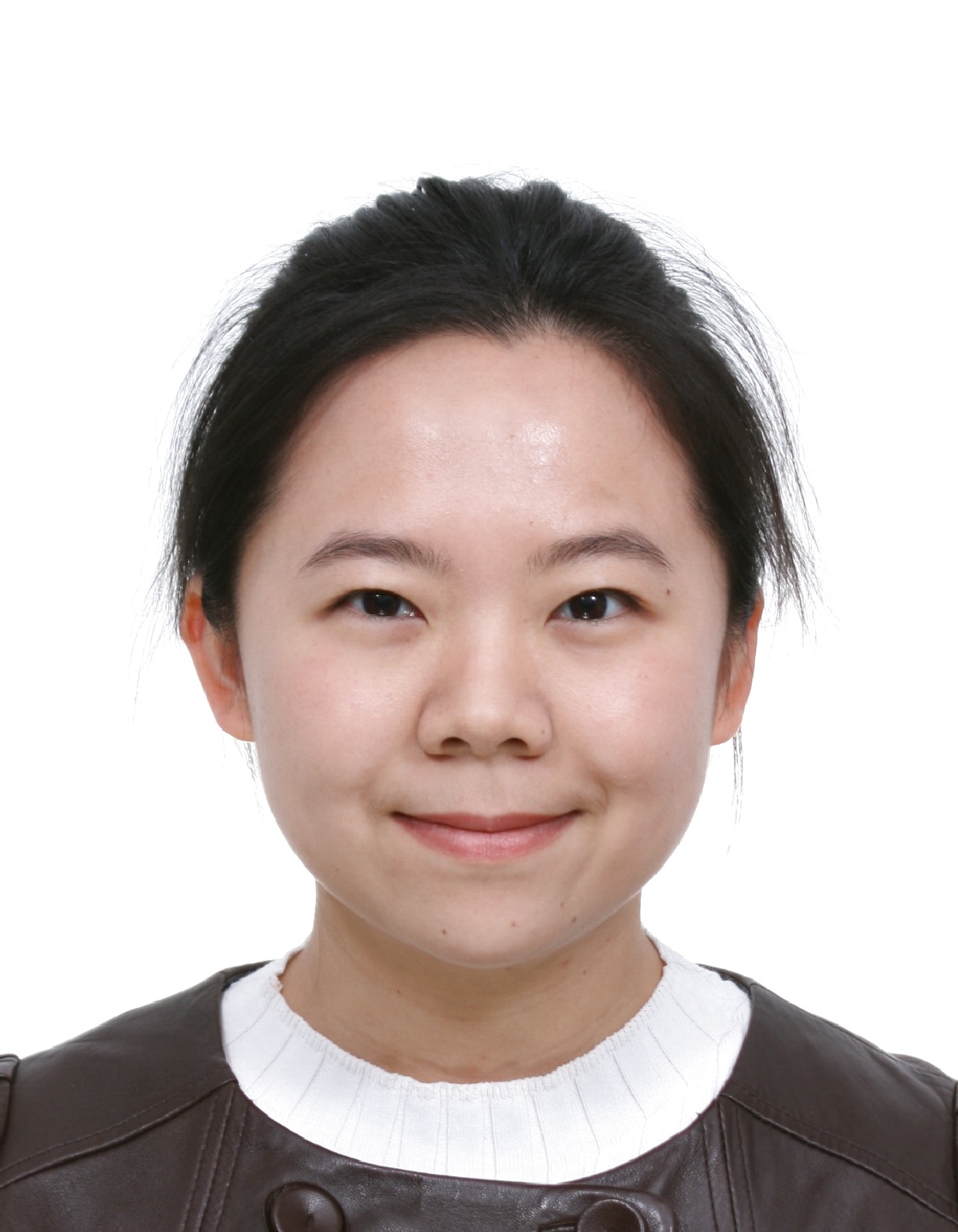}}]{Xiao Pu}
received the Ph.D. degree in electrical engineering from École Polytechnique Fédérale de Lausanne (EPFL), Switzerland in 2018. 
She is currently working at the School of Cyber Security and Information Law in Chongqing University of Posts and Telecommunications.
Her research interests include cross-modal semantic understanding and natural language processing.
\end{IEEEbiography}


\begin{IEEEbiography}[{\includegraphics[width=1in,height=1.25in,clip,keepaspectratio]{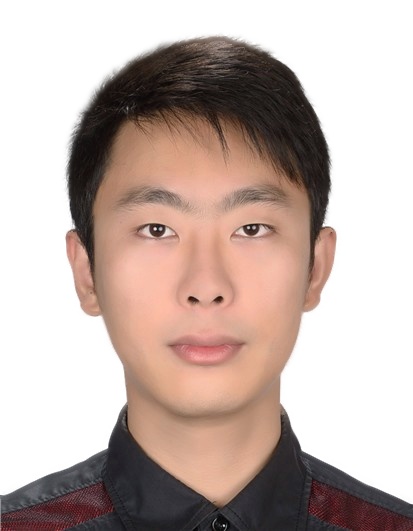}}]{Yan Zhang}
received the B.E. degree from National University of Defense Technology, Changsha, China, in 2014, and the M.S. degree from the University of Birmingham, United Kingdom, in 2017, and the Ph.D. degree from Chongqing University, Chongqing, China, in 2020.
He is currently a lecturer in Chongqing University of Posts and Telecommunications. His research area includes image processing, machine learning, and pattern recognition.
\end{IEEEbiography} 


\begin{IEEEbiography}[{\includegraphics[width=1in,height=1.25in,clip,keepaspectratio]{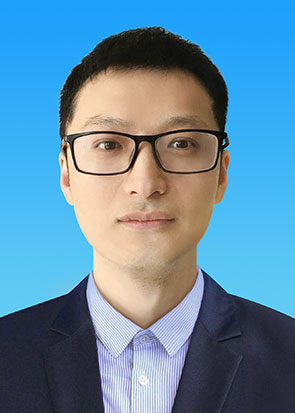}}]{Jiaxu Leng}
Jiaxu Leng received the Ph.D. degree in computer science at University of Chinese Academy of Sciences, Beijing, China, in 2020. He is currently working at School of Computer Science and Technology, Chongqing University of Posts and Telecommunications. His current research interests include computer vision, video anomaly detection, and object detection.
\end{IEEEbiography}


\begin{IEEEbiography}[{\includegraphics[width=1in,height=1.25in,clip,keepaspectratio]{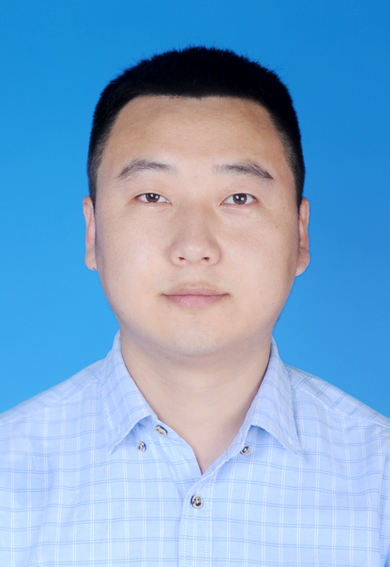}}]{Tao Wu}
is currently a Professor at the School of CyberSecurity and Information Law, Chongqing University of Posts and Telecommunications, China. He is the Executive Deputy Director of Chongqing Network and Information Security Technology Engineering Laboratory. He received the Ph.D. degree from University of Electronic Science and Technology of China, in June 2017. His research interests include graph neural networks, artificial intelligence (AI) security, and graph mining. 
\end{IEEEbiography}


\begin{IEEEbiography}[{\includegraphics[width=1in,height=1.25in,clip,keepaspectratio]{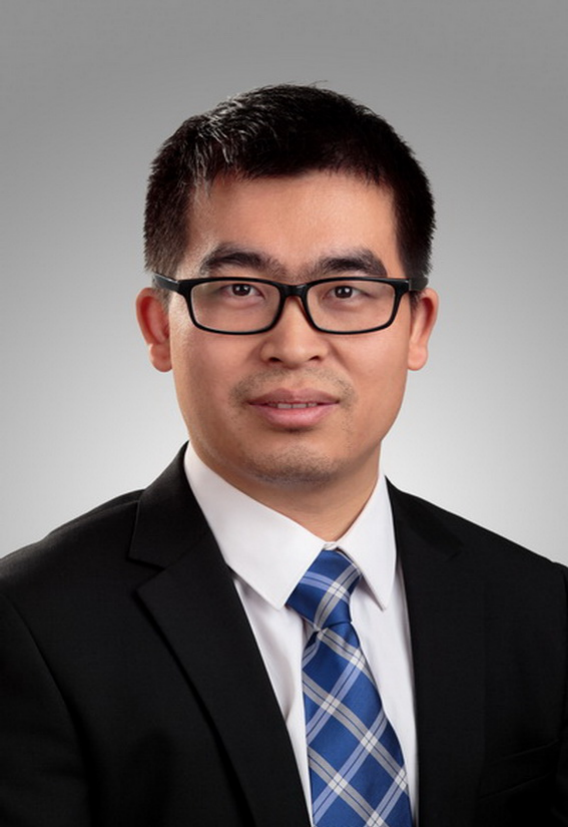}}]{Nannan Wang}
received the B.Sc.
degree in information and computation science from
the Xi’an University of Posts and Telecommunications
in 2009 and the Ph.D. degree in information
and telecommunications engineering from Xidian
University in 2015. He is currently a Professor with
the State Key Laboratory of Integrated Services
Networks, Xidian University, Xi'an, China. He has
published over 150 articles in journals and
proceedings including IEEE T-PAMI, IJCV, CVPR and ICCV. 
His research interests include computer vision and machine learning.
\end{IEEEbiography}


\begin{IEEEbiography}[{\includegraphics[width=1in,height=1.25in,clip,keepaspectratio]{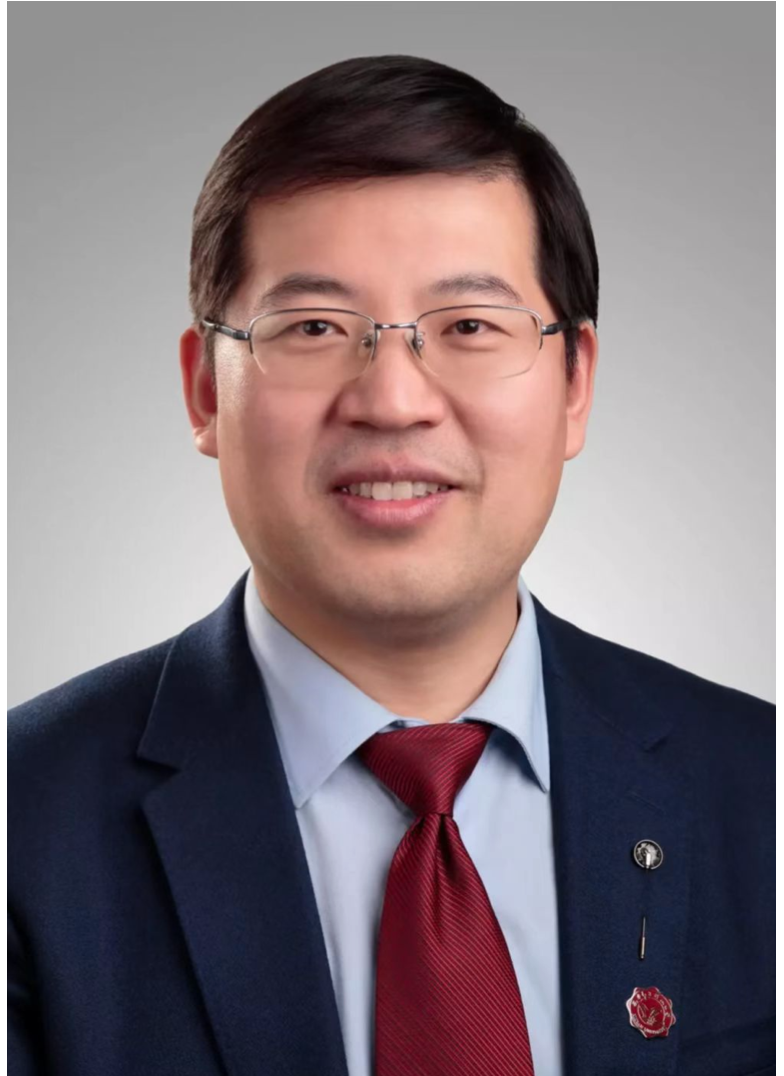}}]{Xinbo Gao}
received the B.Eng., M.Sc., and Ph.D. degrees in signal and information processing from Xidian University, Xi'an, China, in 1994, 1997, and 1999, respectively. From 1997 to 1998, he was a Research Fellow with the Department of Computer Science, Shizuoka University, Shizuoka, Japan. From 2000 to 2001, he was a Post-Doctoral Research Fellow with the Department of Information Engineering, The Chinese University of Hong Kong, Hong Kong. Since 2001, he has been with the School of Electronic Engineering, Xidian University. He is currently a Cheung Kong Professor of Ministry of Education, a Professor of Pattern Recognition
and Intelligent System, and the Director of the State Key Laboratory of Integrated Services Networks, Xi'an. His current research interests include multimedia analysis, computer vision, pattern recognition, machine learning, and wireless communications. 
He has published five books and around 200 technical articles in refereed journals and proceedings. 
\end{IEEEbiography}

\vfill

\end{document}